\documentclass[10pt,twocolumn,letterpaper]{article}

\usepackage[pagenumbers]{cvpr} % To force page numbers, e.g. for an arXiv version
\usepackage{graphicx}
\usepackage{amsmath}
\usepackage{amssymb}
\usepackage{booktabs}
\usepackage{animate}
\usepackage{enumitem}
\usepackage{multirow}
\usepackage{diagbox}

\usepackage{amsmath}
\usepackage{amssymb}
\usepackage{booktabs}
\usepackage{float}
\usepackage{diagbox}
\usepackage{multirow}
\usepackage{enumitem}
\usepackage{tabularx}
\usepackage{microtype}
\usepackage{soul}
\usepackage[accsupp]{axessibility} % Improves PDF readability for those with visual impairments.
\usepackage[dvipsnames]{xcolor}
\definecolor{cvprblue}{rgb}{0.21,0.49,0.74}
\usepackage[pagebackref,breaklinks,colorlinks,citecolor=cvprblue]{hyperref}
\usepackage{tikz}
\def\paperID{2393} % *** Enter the Paper ID here
\def\confName{CVPR}
\def\confYear{2024}

\title{Drag Your Noise: Interactive Point-based Editing \\via Diffusion Semantic Propagation\vspace{-5mm}}

\author{Haofeng Liu\textsuperscript{1,2\textsuperscript{\textasteriskcentered}}
\and Chenshu Xu\textsuperscript{1\thanks{The first two authors contributed equally.}}
\and Yifei Yang\textsuperscript{1}
\and Lihua Zeng\textsuperscript{2}
\and Shengfeng He\textsuperscript{1\thanks{Corresponding author (\emph{shengfenghe@smu.edu.sg}).}}
\and \textsuperscript{1}Singapore Management University
\and \textsuperscript{2}South China Normal University
}

\begin{document}

\teaser{%\vspace{5mm}
    \centering
    \begin{subfigure}{.12\linewidth}
        \centering
        \includegraphics[width=\linewidth]{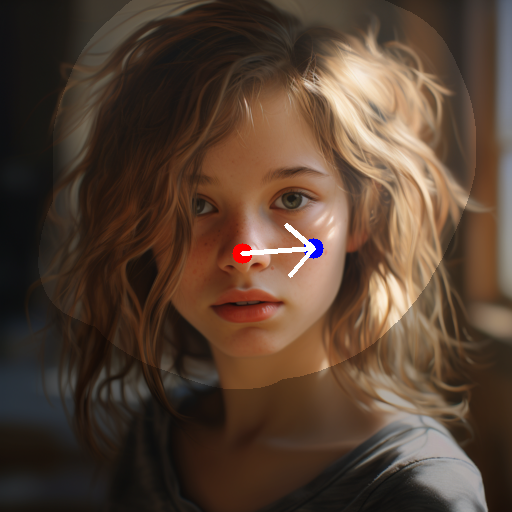}
        \includegraphics[width=\linewidth]{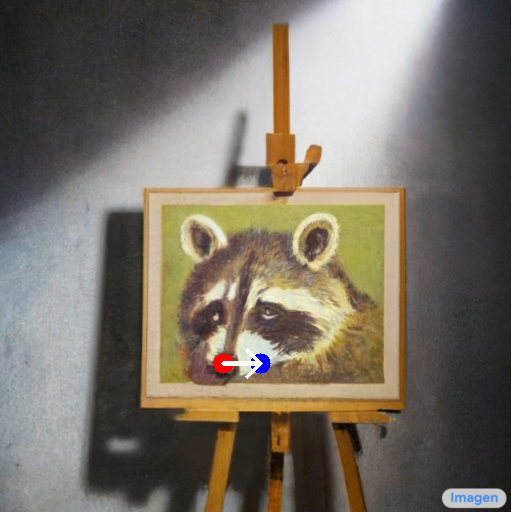}
    \caption{\footnotesize{Input}\label{fig_lta}}
    \end{subfigure}
    \begin{subfigure}{.12\linewidth}
        \centering
        \includegraphics[width=\linewidth]{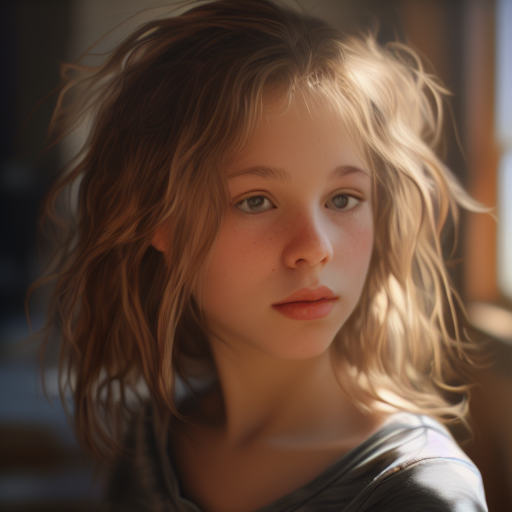}
        \includegraphics[width=\linewidth]{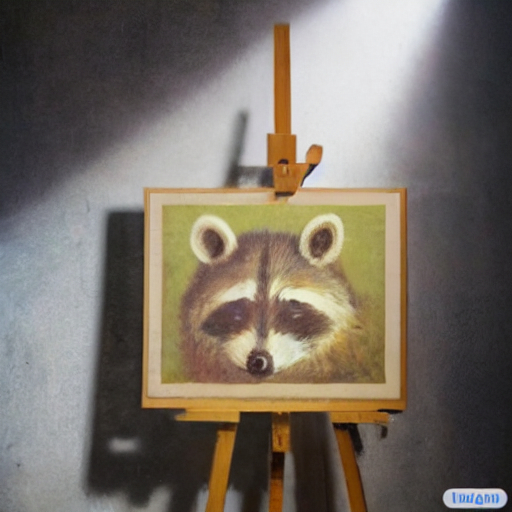}
    \caption{\footnotesize{DragNoise}\label{fig_ltb}}
    \end{subfigure}
    \begin{subfigure}{.12\linewidth}
        \centering
        \includegraphics[width=\linewidth]{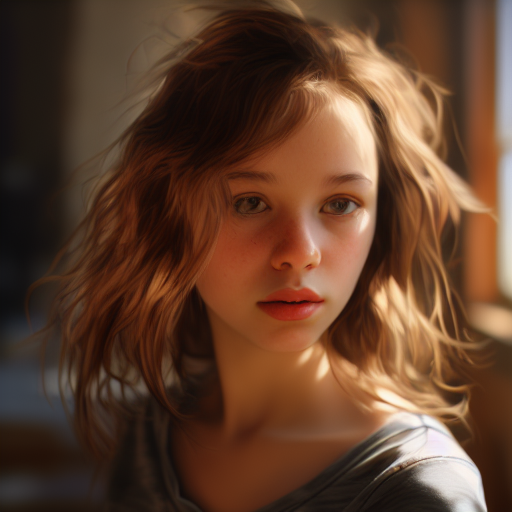}
        \includegraphics[width=\linewidth]{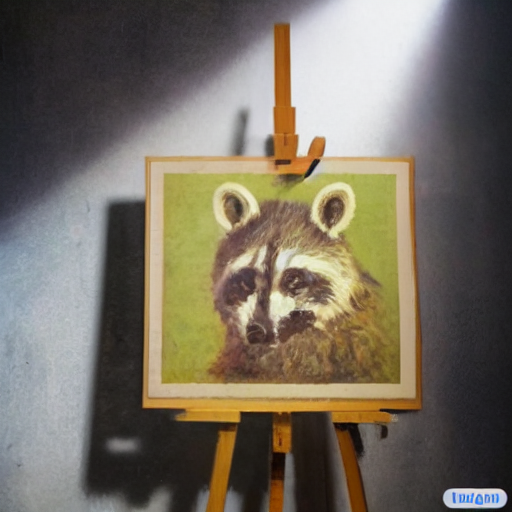}
    \caption{\footnotesize{DragDiffusion}\label{fig_ltc}}
    \end{subfigure}
    \begin{subfigure}{.12\linewidth}
        \centering
        \includegraphics[width=\linewidth]{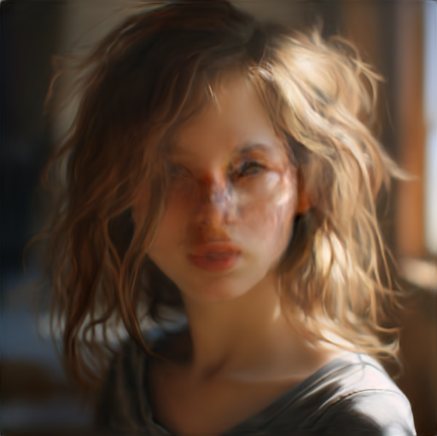}
        \includegraphics[width=\linewidth]{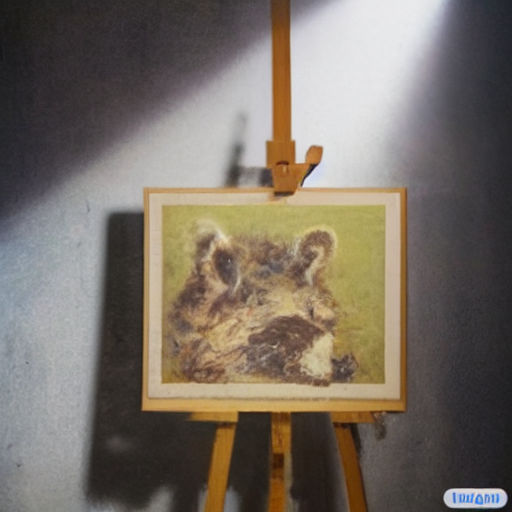}
    \caption{\footnotesize{DragGAN}\label{fig_ltd}}
    \end{subfigure}
    \begin{subfigure}{.12\linewidth}
        \centering
        \includegraphics[width=\linewidth]{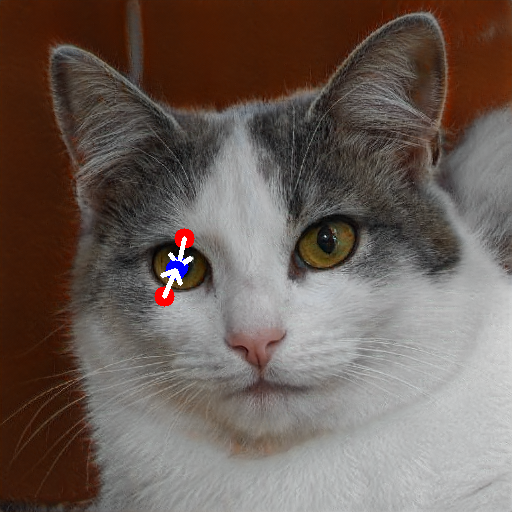}
        \includegraphics[width=\linewidth]{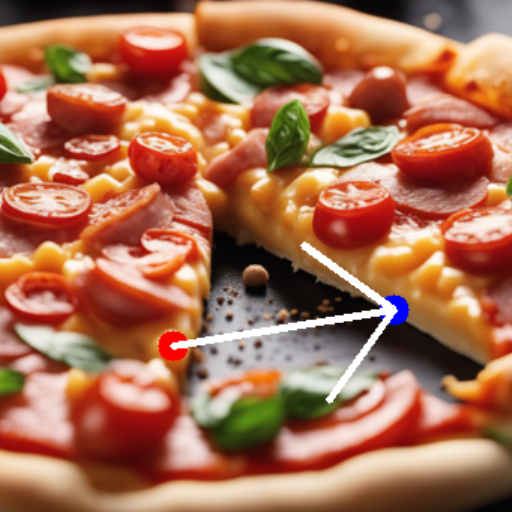}
    \caption{\footnotesize{Input}\label{fig_lte}}
    \end{subfigure}
    \begin{subfigure}{.12\linewidth}
        \centering
        \includegraphics[width=\linewidth]{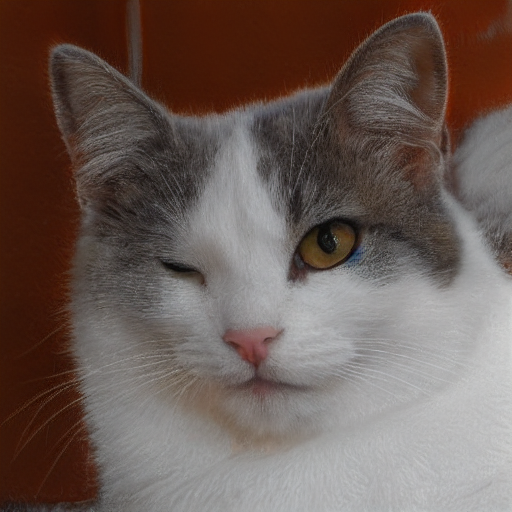}
        \includegraphics[width=\linewidth]{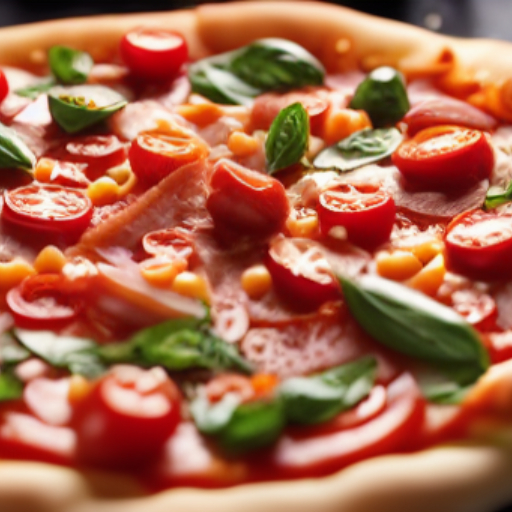}
    \caption{\footnotesize{DragNoise}\label{fig_ltf}}
    \end{subfigure}
    \begin{subfigure}{.12\linewidth}
        \centering
        \includegraphics[width=\linewidth]{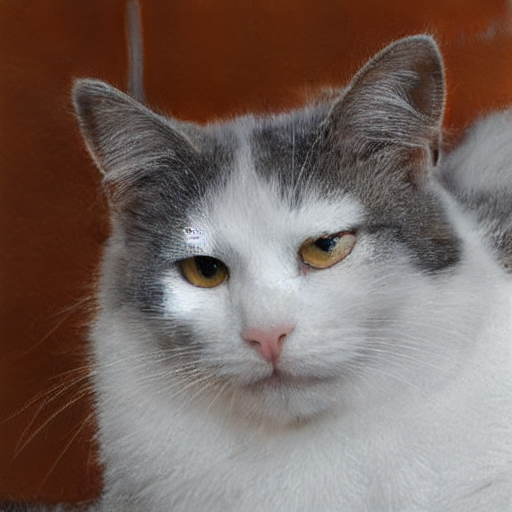}
        \includegraphics[width=\linewidth]{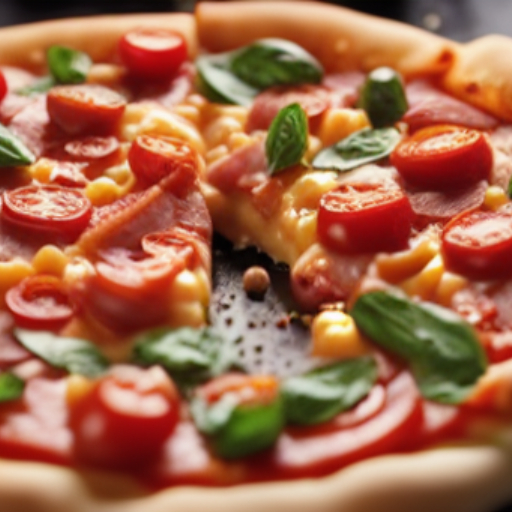}
    \caption{\footnotesize{DragDiffusion}\label{fig_ltg}}
    \end{subfigure}
    \begin{subfigure}{.12\linewidth}
        \centering
        \includegraphics[width=\linewidth]{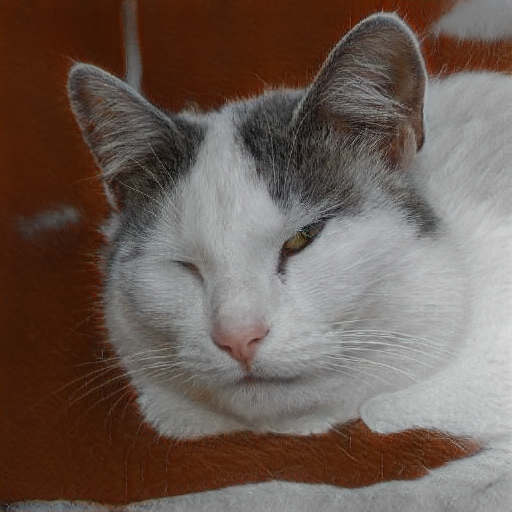}
        \includegraphics[width=\linewidth]{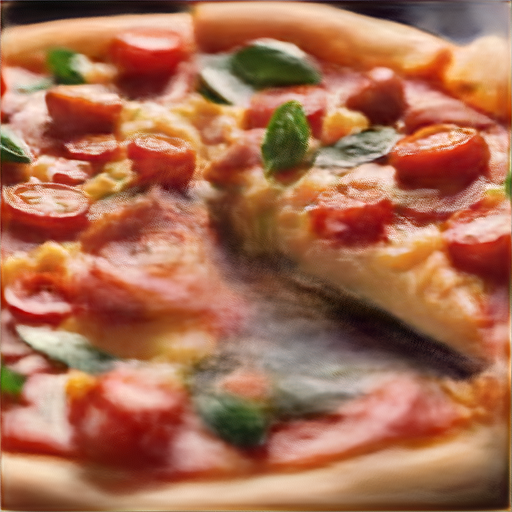}
    \caption{\footnotesize{DragGAN}\label{fig_lth}}
    \end{subfigure}
\vspace{-3mm}\caption{We present DragNoise, a point-based interactive editing framework that avoids the global adjustments of the latent code/map, a common issue in frameworks like DragGAN~\cite{pan2023drag} and DragDiffusion~\cite{shi2023dragdiffusion}, facilitating stable and semantically accurate point-based editing.}
\vspace{-5mm}
\label{fig:teaser}
}

\maketitle

\begin{abstract}\vspace{-4mm}
Point-based interactive editing serves as an essential tool to complement the controllability of existing generative models. A concurrent work, DragDiffusion, updates the diffusion latent map in response to user inputs, causing global latent map alterations. This results in imprecise preservation of the original content and unsuccessful editing due to gradient vanishing. In contrast, we present DragNoise, offering robust and accelerated editing without retracing the latent map. The core rationale of DragNoise lies in utilizing the predicted noise output of each U-Net as a semantic editor. This approach is grounded in two critical observations: firstly, the bottleneck features of U-Net inherently possess semantically rich features ideal for interactive editing; secondly, high-level semantics, established early in the denoising process, show minimal variation in subsequent stages. Leveraging these insights, DragNoise edits diffusion semantics in a single denoising step and efficiently propagates these changes, ensuring stability and efficiency in diffusion editing. Comparative experiments reveal that DragNoise achieves superior control and semantic retention, reducing the optimization time by over 50\% compared to DragDiffusion. Our codes are available at \href{https://github.com/haofengl/DragNoise}{https://github.com/haofengl/DragNoise}.\vspace{-5mm}
\end{abstract}

\section{Introduction}
\label{sec:intro}
The limited controllability inherent in diffusion models~\cite{ho2020denoising,dhariwal2021diffusion} highlights the need for interactive editing in image manipulation. Consequently, recent advancements have led to a variety of interactive approaches. These include text-guided editing~\cite{kawar2023imagic,hertz2022prompt,avrahami2022blended,brooks2023instructpix2pix,li2023blipdiffusion}, stroke-based editing~\cite{meng2021sdedit}, and exemplar-based methods~\cite{yang2023paint,gu2023photoswap,jiang2023diffuse3d,Yu2024Beyond}. As the demand for more user-friendly and precise editing methods grows, the implementation of drag-and-drop manipulation of control points has emerged as a straightforward and efficient approach in real-world applications.

In the field of point-based image editing, DragGAN~\cite{pan2023drag} represents a significant milestone by leveraging generative adversarial networks (GANs)~\cite{goodfellow2014generative,karras2020analyzing}. Despite its innovation, the inherent constraints of GANs often limit achieving high-quality edited outcomes. Additionally, GAN-based editing methods~\cite{xia2022gan,zhu2020domain,abdal2019image2stylegan,xu2023rigid,zheng2024learning}, which involve optimizing a new latent code corresponding to the edited result, struggle with preserving global content, as illustrated in \cref{fig_ltd,fig_lth}. Different from traditional ``outer-inversion'' that converts a real image into a latent code, we term this internal optimization process of inverting user editing into the latent code as ``inner-inversion'' (see \cref{fig:draggan}).
A concurrent work, DragDiffusion~\cite{shi2023dragdiffusion} progresses this field by applying diffusion models, capitalizing on the strengths of large-scale pre-trained models. Although DragDiffusion applies diffusion models, it adheres to the concept of ``inner-inversion''. This method guides the optimization of intermediate noisy latent maps to generate outputs that reflect the intended editing (\cref{fig:dragdiffusion}).

However, two main issues emerge with DragDiffusion: \textbf{gradient vanishing} and \textbf{inversion fidelity}. Gradient vanishing during optimization occurs due to the reliance on motion supervision loss, which is based on feature differences before and after dragging. This issue is exacerbated when feature differences are minimal and the back-propagation chain in inversion is lengthy, leading to ``under-dragging'' in their results, as evident in the bottom of \cref{fig_ltc,fig_ltg}. Additionally, maintaining reconstruction fidelity remains a longstanding challenge in inversion techniques. Although DragDiffusion~\cite{shi2023dragdiffusion} improves spatial control by extending ``inner-inversion'' to 2D latent maps, surpassing DragGAN's optimization of 1D latent codes, it still struggles in fine reconstruction {due to its optimization path back to the noisy latent map}, as demonstrated in the top of \cref{fig_ltc,fig_ltg}.

\begin{figure}
    \rotatebox[origin=c]{270}{\hspace{-77.5mm} \small{$t = 25$} \hspace{-27.5mm} \small{$t = 35$} \hspace{-27.5mm} \small{$t = 45$} \hspace{-29mm} \small{Original}}
    \centering
    \begin{subfigure}{.31\linewidth}
    \centering
        \includegraphics[width=1.0\textwidth]{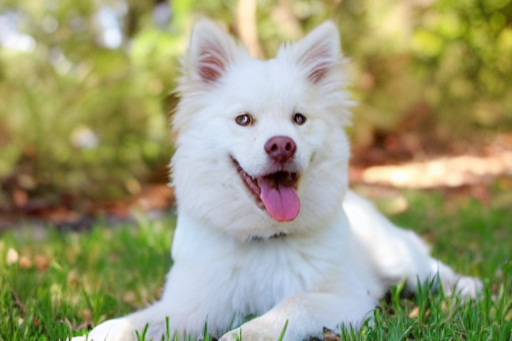}\\
        \includegraphics[width=1.0\textwidth]{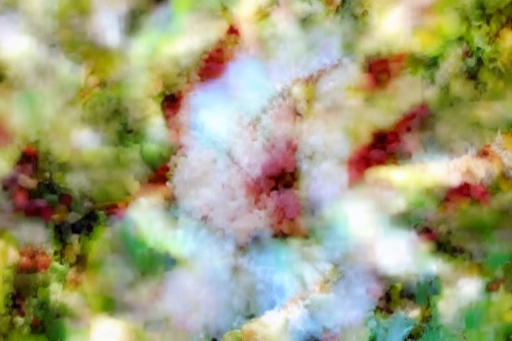}\\
        \includegraphics[width=1.0\textwidth]{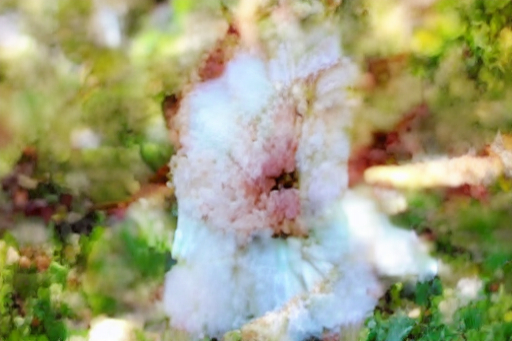}\\
        \includegraphics[width=1.0\textwidth]{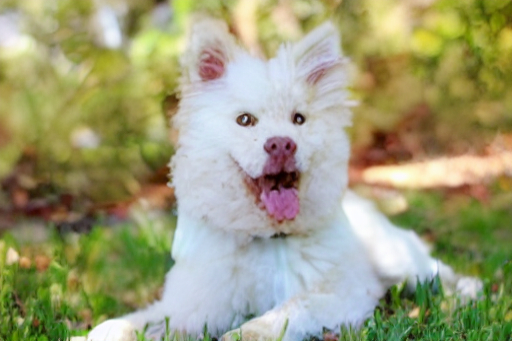}
    \caption{\footnotesize{Encoder Block 1}\label{fig_encoder1}}
    \end{subfigure}
    % \hspace{-1.5mm}
    \begin{subfigure}{.31\linewidth}
        \centering
        \includegraphics[width=1.0\textwidth]{image/inversion.jpg}\\
        \includegraphics[width=1.0\textwidth]{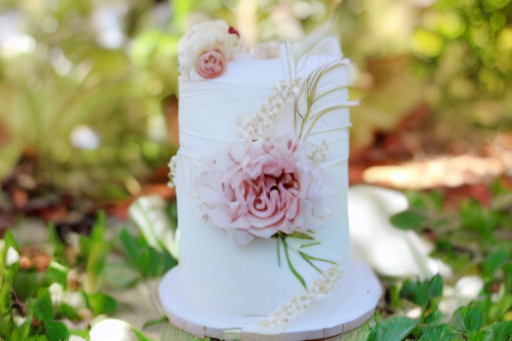}\\
        \includegraphics[width=1.0\textwidth]{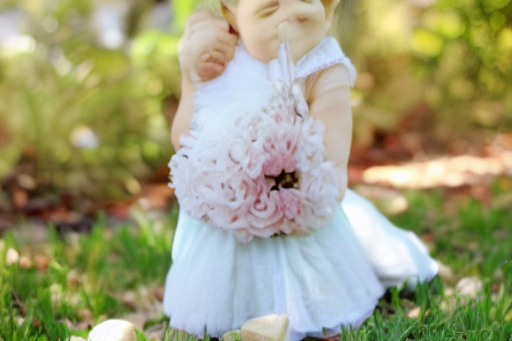}\\
        \includegraphics[width=1.0\textwidth]{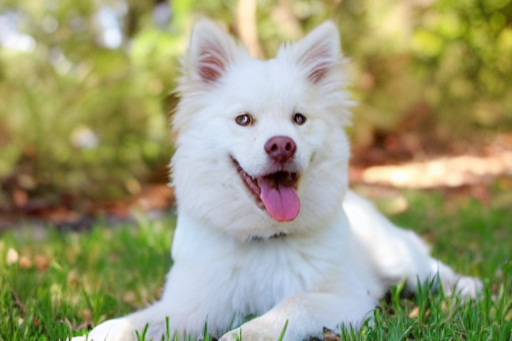}
    \caption{\footnotesize{Encoder Block 3}\label{fig_encoder3}}
    \end{subfigure}
    % \hspace{-1.5mm}
    \begin{subfigure}{.31\linewidth}
        \centering
        \includegraphics[width=1.0\textwidth]{image/inversion.jpg}\\
        \includegraphics[width=1.0\textwidth]{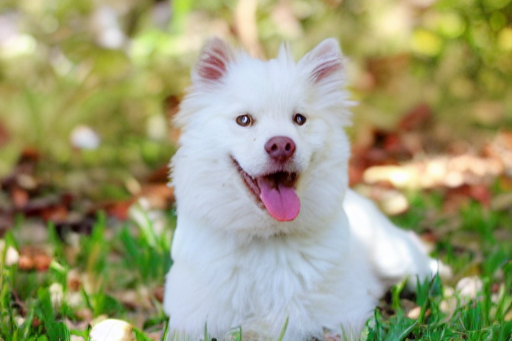}\\
        \includegraphics[width=1.0\textwidth]{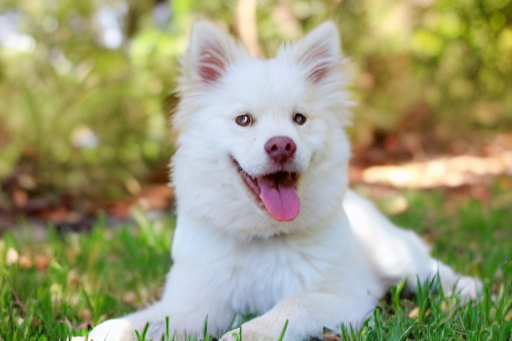}\\
        \includegraphics[width=1.0\textwidth]{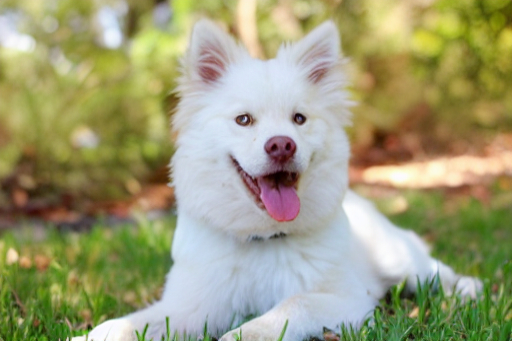}
    \caption{\footnotesize{Bottleneck}\label{fig_bottleneck}}
    \end{subfigure}
    % \hspace{-1.5mm}
    \vspace{-7mm}\caption{Reconstructed images by DDIM inversion, where features of different levels (column) are copied to corresponding layers in all subsequent U-Nets, beginning from various denoising timesteps (row). The original images, reconstructed without feature copying, are provided for comparison. This synchronization of bottleneck features across all subsequent steps reveals that the core semantics of the diffusion process are encoded within the bottleneck layer, predominantly learned in the early phases of the denoising process.}
    \vspace{-5mm}
    \label{fig:copy}
\end{figure}

\begin{figure}
    \centering
    \subfloat{\includegraphics[width=0.49\textwidth]{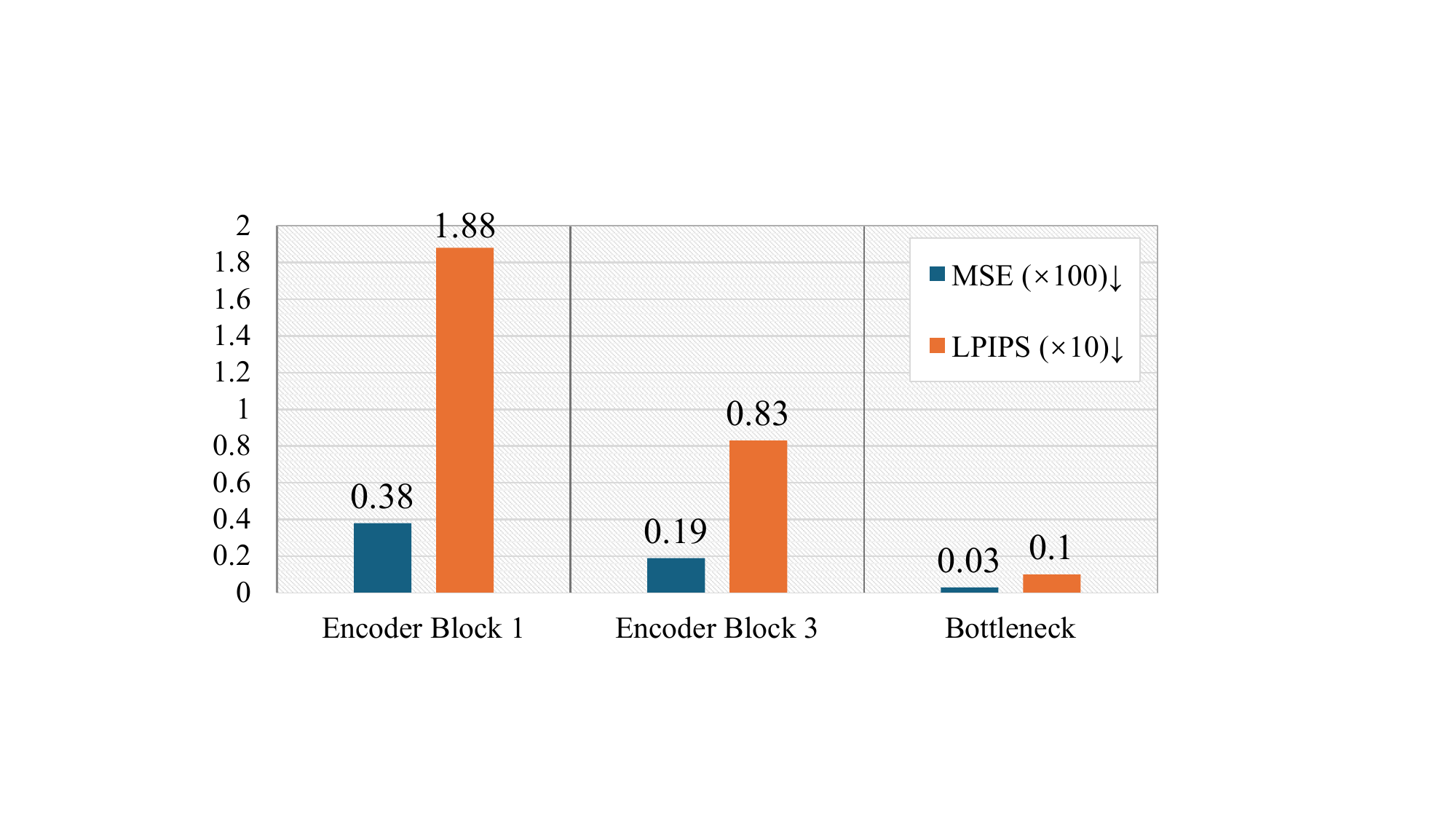}}
    \vspace{-3mm}\caption{Quantitative analysis on middle-block feature replacement. This involves replacing features at all subsequent timesteps with those from timestep 35, using features from various layers. Our evaluation metrics, MSE and LPIPS~\cite{zhang2018unreasonable} were used to compare the reconstructed images against the original inputs.}\vspace{-5mm}
    \label{fig:numerical}
 \end{figure}

\noindent\textbf{Diffusion Semantic Analysis.} Prior studies~\cite{baranchuk2022label,voynov2023p+} have demonstrated that intermediate diffusion features from the noise predictor not only effectively capture semantic information but also facilitate structure-to-appearance controllability. Inspired by them, we reevaluate the necessity of retracing the latent map and explore diffusion semantics within the editing mechanism. Initially, we conduct feature analysis using DDIM inversion~\cite{dhariwal2021diffusion,song2020denoising} on the pre-trained Stable Diffusion~\cite{rombach2022high} model. To understand the semantics learned by different U-Net layers, we copy features from these layers (shown in different columns) and replace the corresponding ones in all subsequent U-Nets, starting from various denoising timesteps (shown in different rows). This is to show \emph{where} and \emph{when} diffusion models learn semantic knowledge. {The resulting images and numerical results are shown in \cref{fig:copy} and \cref{fig:numerical} respectively.}

As the U-Net encoder progressively shrinks the features, higher-level features are attained as the network approaches the bottleneck. The reconstructed images demonstrate that the substitution of low-level features (``Encoder Block 1'' and ``Encoder Block 3'') compromises image reconstruction detail and quality. This outcome is attributed to the distinct roles of low-level features at different timesteps that cannot be shared, such as adding fine textures in the final stages. On the other hand, the bottleneck, producing the high-level feature in the noise predictor, exhibits the capacity to capture more complete semantics even during early timesteps. Generally, copying and replacing features at later timesteps yield better reconstruction of the original image. However, at an early timestep 45, the bottleneck feature can encapsulate the rough outline of the dog, albeit missing finer details like ears and legs. Interestingly, substituting the bottleneck feature from timestep 35 preserves the overall structure, and propagating this early timestep semantic to subsequent steps does not diminish reconstruction quality. These findings lead us to conclude that the bottleneck feature represents an optimal diffusion semantic representation, particularly suitable for efficient editing. Since it can be effectively trained in early timesteps, manipulating bottleneck features allows for smooth propagation to later denoising steps, ensuring the integrity of complete diffusion semantics is maintained. Moreover, due to the short optimization path, the problem of gradient vanishing is efficiently avoided.

\noindent\textbf{Noise Maps as Semantic Editors.} Building on our previous analysis, we introduce DragNoise, an interactive point-based image editing method that leverages diffusion semantic propagation. The rationale behind our DragNoise is to treat the predicted noises as sequential semantic editors. Our editing process initiates at a timestep (\eg, t=35) where high-level semantics are well-trained. During this phase, diffusion semantic optimization is conducted on the bottleneck feature of the U-Net to reflect user edits. The optimized bottleneck feature learns the intended dragging effect and produces the corresponding manipulation noise. This optimized bottleneck feature contains target semantics and therefore is propagated to all subsequent timesteps by substituting the corresponding bottleneck features, avoiding redundant feature optimization. This substitution significantly augments the manipulation effect in a stable and efficient manner.
We conducted extensive quantitative and qualitative experiments on the drag-based editing benchmark DragBench~\cite{shi2023dragdiffusion} and diverse example images to evaluate the efficacy of our DragNoise. Notably, DragNoise significantly cuts down the optimization time by over 50\% compared to DragDiffusion. Additionally, we explore the impacts of various initial timesteps on the editing process, the optimization of different layers, and the extent of optimization propagation, which underscore DragNoise's efficiency and flexibility in interactive editing. 
\section{Related Work}
\label{sec:related}

\textbf{Image Editing.}
The field of image editing has advanced remarkably, paralleling improvements in image synthesis quality~\cite{karras2019style,karras2020analyzing,dhariwal2021diffusion,rombach2022high}. Building on StyleGAN~\cite{karras2019style,karras2020analyzing}, numerous studies leverage the latent space of pre-trained GANs for diverse image manipulations, including recoloring~\cite{afifi2021histogan}, semantic or attribute adjustments~\cite{patashnik2021styleclip,parihar2022everything,shen2020interpreting,yang2021discovering,xu2021continuity,song2022editing}, and style transfer~\cite{yang2022pastiche,wu2022make,li2023parsing}.

With the advancement of diffusion models, many researchers have applied diffusion models to similar editing tasks. SDEdit~\cite{meng2021sdedit} represents an early exploration of semantics in the denoising process, editing images via pixel-level guidance through noise addition and denoising. However, it is primarily limited to global editing. As large-scale pre-trained latent diffusion models (LDMs)~\cite{rombach2022high} emerge as frontrunners in generative modeling, their principles are increasingly utilized in various editing tasks. For localized editing, some studies~\cite{hertz2022prompt,gu2023photoswap} utilize cross-attention mechanisms. Additionally, given that latent maps are intermediate image representations, they are apt for image editing~\cite{choi2021ilvr}. However, unlike the latent space in StyleGANs, directly editing the latent maps in diffusion models performs poorly in the presence of significant content or color changes. While most diffusion model editing tasks rely on text prompts~\cite{kawar2023imagic,avrahami2022blended,brooks2023instructpix2pix,li2023blipdiffusion,mokady2023null}, there is a growing need for more convenient, precise, and user-friendly interactive control methods for users.

%-------------------------------------------------------------------------

\noindent\textbf{Point-based Interactive Editing.}
Point-based image editing, known for its user-friendliness, enables a wide range of effects from fine-grained adjustments to more extensive transformations. In prior works, this typically involves a two-step optimization in the latent space: motion supervision and point tracking. A notable example utilizing this principle is DragGAN~\cite{pan2023drag}. However, the generative limitations of GANs restrict more complex edits, particularly with real images. FreeDrag~\cite{ling2023freedrag}, another GAN-based approach, adopts a feature-oriented strategy to diminish dependence on point tracking. Yet, it still faces constraints due to GANs' limitations in generative capabilities, inversion efficiency, and the semantic versatility of latent codes. Recently, DragDiffusion~\cite{shi2023dragdiffusion} has adapted DragGAN's principles to diffusion models, resulting in improved spatial control and greater flexibility. In a similar vein, DragonDiffusion~\cite{mou2023dragondiffusion} employs classifier guidance and translates editing signals into gradients through feature correspondence loss. Diverging from these methods, our approach exploits diffusion semantics within diffusion models for point-based editing, offering a novel perspective in interactive image manipulation.

\section{Drag Your Noise}
\label{sec:method}
\begin{figure*}
  \begin{minipage}[b]{0.26\textwidth}
    \begin{subfigure}[b]{\textwidth}
      \includegraphics[width=1.0\linewidth]{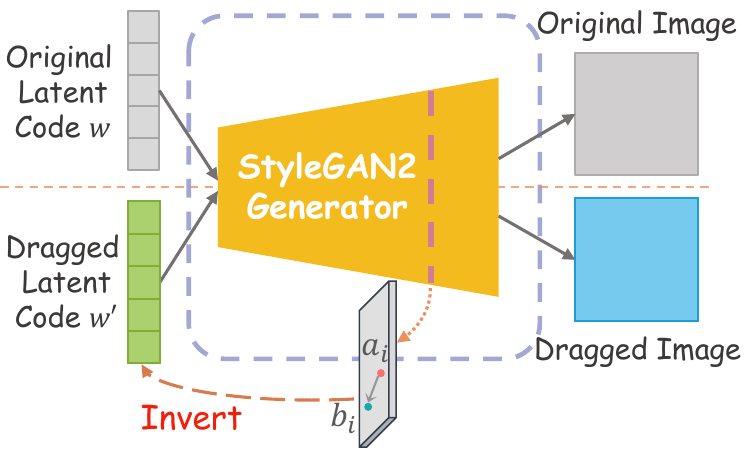}
      \caption{DragGAN}
      \label{fig:draggan}
    \end{subfigure}
    \begin{subfigure}[b]{\textwidth}
      \includegraphics[width=1.0\linewidth]{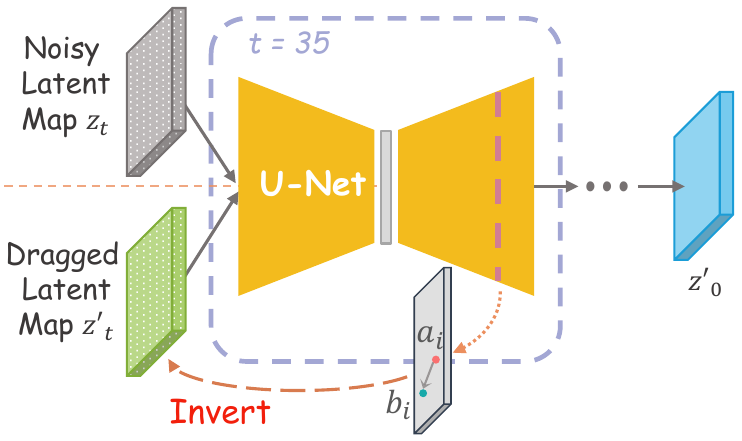}
      \caption{DragDiffusion}
      \label{fig:dragdiffusion}
    \end{subfigure}
  \end{minipage}%
  \hspace{3mm}
  \tikz{\draw[-,gray, densely dashed, thick](0,-0.55) -- (0,6.5);}
  \hspace{2mm}
  \begin{subfigure}[b]{0.69\textwidth}
    \includegraphics[width=1.0\linewidth]{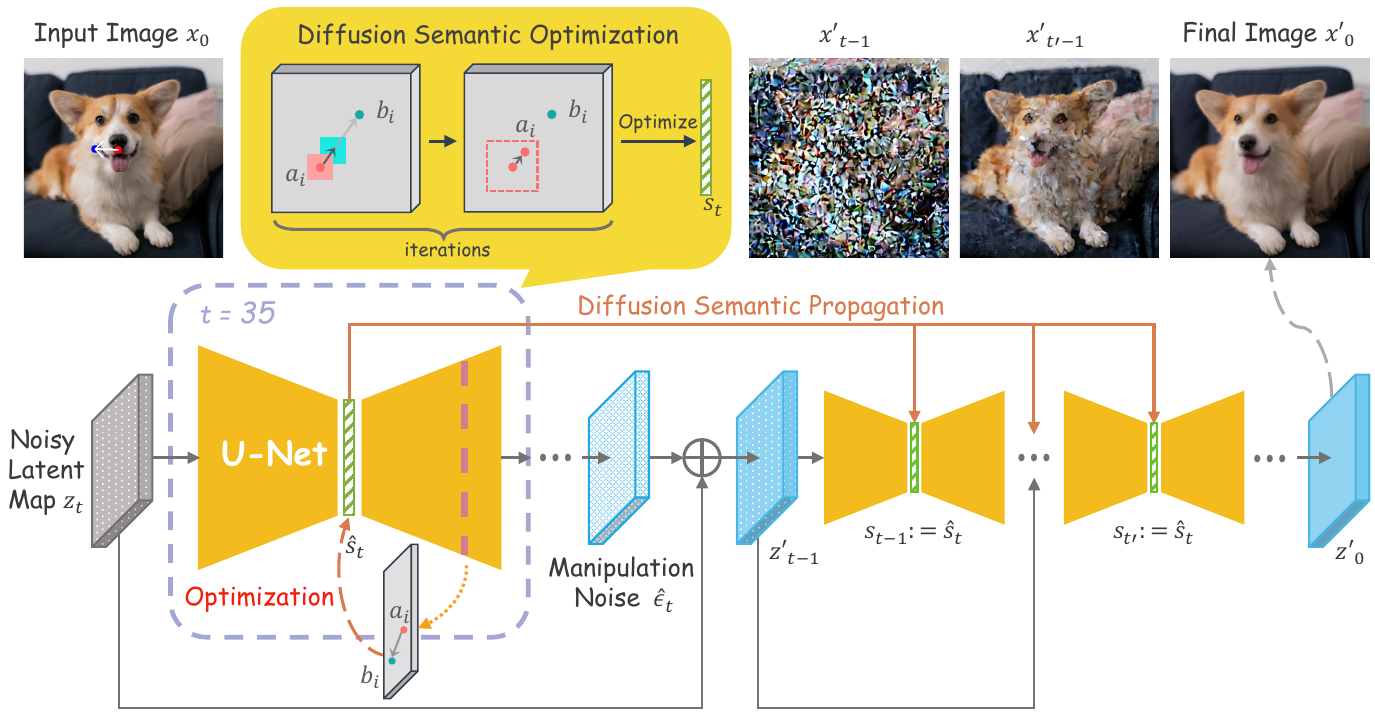}
    \caption{DragNoise}
    \label{fig:dragnoise}
  \end{subfigure}
  \vspace{-3mm}\caption{Comparison between DragNoise and other relevant methods in feature modification, highlighting optimized features in green.}\vspace{-5mm}
  \label{fig:overview}
\end{figure*}

In this section, we present our method, DragNoise, designed for interactive point-based editing. To foster a comprehensive understanding, we begin by revisiting the fundamentals of diffusion models and LDMs. Additionally, we conduct a concise comparative analysis with previous methodologies. Distinguishing DragNoise from earlier approaches, our method focuses on manipulating the predicted noise, encompassing diffusion semantic optimization and diffusion semantic propagation.

\subsection{Preliminaries}
Diffusion models~\cite{sohl2015deep,ho2020denoising} are probabilistic models that learn a data distribution by iteratively denoising noisy images. The forward process refers to a Markov process where the data is corrupted by Gaussian noise.
Diffusion models are trained to approximate the reverse process. Practically, a network, typically implemented with U-Net~\cite{ronneberger2015u}, is used to predict the noise $\hat \epsilon_t$ at timestep $t$, denoted as $\epsilon_\theta(x_t, t); t = 1,2, ..., T$.
% \todo{The characteristic of U-Net, shape of U}
In each timestep, the U-Net receives a noisy image as its input and generates predictions for the noise component at this timestep.
Recently, there has been a notable emergence of large-scale diffusion models showing impressive generative capability, especially those based on LDMs, upon which our method is implemented. LDMs compress the denoising process onto the latent space. Given an input noisy latent map $z_t$, the denoising U-Net in LDMs is denoted as $\epsilon_\theta(z_t, t)$.

{To enable point-based editing, DragDiffusion extends the inner-inversion method of DragGAN with LDMs (see \cref{fig:draggan,fig:dragdiffusion}).}
DragGAN employs feature maps after the 6th block of the StyleGAN2~\cite{karras2020analyzing} generator for motion supervision, inverting the editing to the dragged latent code $w'$, which produces the corresponding dragged image. Similarly, DragDiffusion utilizes a U-Net feature map to supervise the optimization of the noisy latent map, yielding a dragged latent map $z'_t$ that governs the subsequent denoising process.
Different from their approaches, we manipulate predicted noise and propagate the optimization to edit the image.

\subsection{Methodology}

\subsubsection{Diffusion Semantic Optimization}
We present an overview of our methodology in \cref{fig:dragnoise}.
% During the diffusion generation process, we denoted the initial latent map as $z_T$.
Based on the analysis in \cref{sec:intro}, the bottleneck feature of U-Net in diffusion models has the capability of capturing more comprehensive noise semantics compared to other features. Moreover, the bottleneck feature efficiently captures the majority of semantics at a certain early timestep, denoted as $t$. Therefore, given user-defined point instructions, we perform diffusion semantic optimization with the bottleneck feature at this timestep.
% i.e., manipulating the predicted latent noise.
Specifically, inspired by Pan \etal~\cite{pan2023drag}, we perform the optimization process described below.

We denote the user-provided anchor points as $\{a_i=(x^a_i, y^a_i)\}_{i=1,2,...,m}$, and the corresponding objective points as $\{b_i=(x^b_i, y^b_i)\}_{i=1,2,...,m}$, where $m$ is the number of point pairs.
{Direct spatial mapping between points and the bottleneck feature faces challenges due to significant differences in abstraction levels. To overcome this, we introduce an additional intermediate feature. As demonstrated by Baranchuk et al. \cite{baranchuk2022label}, the features from the U-Net decoder's third layer are both informative and aware of structure. Therefore, we utilize these features for supervision, achieving an equilibrium between comprehensive structural representation and granularity of detail.} Specifically, we obtain a point $p$'s corresponding feature element in this feature map via bilinear interpolation, denoted as $F_p$. To align the structural feature of the anchor point to the objective point, we define the semantic alignment loss as:
\begin{equation}
\mathcal{L}_{alignment} = \sum_{i=1}^m \sum_{p_i \in \Omega(a_i, r_1)} \|F_{p_i} - F_{p_i+v_i}\|_1,
\label{eq:loss_align}
\end{equation}
where $v_i=\frac{b_i-a_i}{\|b_i-a_i\|_2}$ is the normalized vector pointing from $a_i$ to $b_i$ and $\Omega(a_i, r_1)=\{(x,y)| |x-x^a_i|\leqslant r_1, |y-y^a_i|\leqslant r_1\}$ denotes the neighborhood of the anchor point. The semantic alignment loss ``drags'' the feature of points near the anchor point towards those near the objective point by a small step.
Moreover, if a mask is provided, we employ a semantic masking loss to keep the bottleneck feature outside the mask unchanged:
\begin{equation}
    \mathcal{L}_{mask} = \|(s_t-\hat s_t) \odot (1-M) \|_1,
\label{eq:loss_mask}
\end{equation}
where $s_t$ is the bottleneck feature at denoising timestep $t$ and $\hat s_t$ is its optimized feature. It is noteworthy that, due to the enhanced semantic decoupling of the bottleneck feature, the mask is generally unnecessary for the majority of cases.

In every iteration, we update the bottleneck feature to influence the generation of manipulation noise with those loss terms above. In total, our optimization goal is defined as:
\begin{equation}
  \hat s_t=\mathop{\arg\min}_{s_t} \ \ ( \mathcal{L}_{alignment}+ \lambda \mathcal{L}_{mask} ).
  \label{eq:objective}
\end{equation}
Note that during back-propagation, the gradient does not propagate backward through $F_{p_i}$.

After each iteration, the bottleneck feature as well as the corresponding feature element of the anchor point are updated. As \cref{eq:loss_align} no longer provides the accurate direction for optimization iterations afterward, we need to update the location of the anchor point after each iteration. This is achieved by searching in $a_i$'s neighborhood $\Omega(a_i, r_2)=\{(x,y)| |x-x^a_i|\leqslant r_2, |y-y^a_i|\leqslant r_2\}$ for the nearest neighbor of the optimized feature:
\begin{equation}
  a_i := \mathop{\arg\min}_{p_i \in \Omega(a_i, r_2)} \ \ \|F'_{p_i}-f^0_i\|_1,
  \label{eq:tracking}
\end{equation}
where $F'_{p_i}$ is feature element of $p_i$ in the updated feature map, and $f^0_i$ denotes the feature element of the initial anchor point.

Finally, diffusion semantic optimization terminates when all the distances between anchor points and their respective objective points are within 1 pixel or when the maximum number of iterations is reached. With the optimized bottleneck feature $\hat s_t$, the decoder of the U-Net produces a manipulation noise $\hat \epsilon_\theta$.

\begin{figure*}
    \captionsetup[subfloat]{position=top}
    % \quad
    \subfloat[Input]{\includegraphics[width=0.19\textwidth]{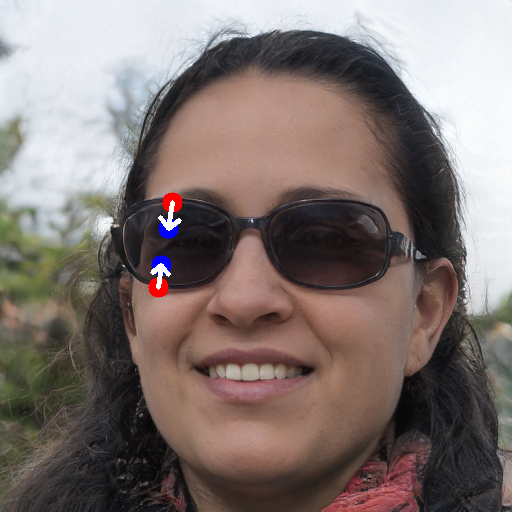}}
    % \hfil
    \hspace{2mm}
    \subfloat[DragNoise]{\label{fig:vsgan-dragnoise}  \includegraphics[width=0.19\textwidth]{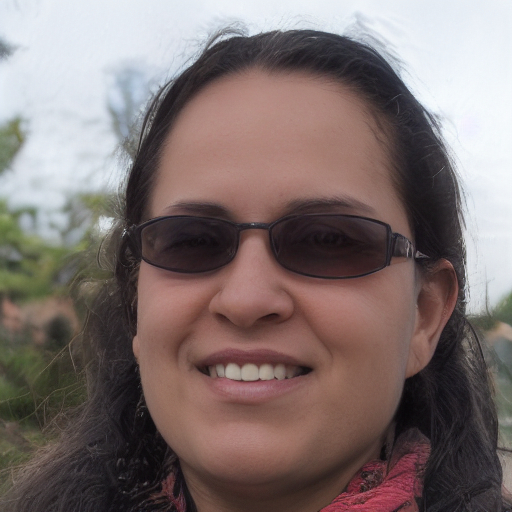}}
    % \hfil
    \subfloat[DragGAN]{\label{fig:vsgan-draggan} \includegraphics[width=0.19\textwidth]{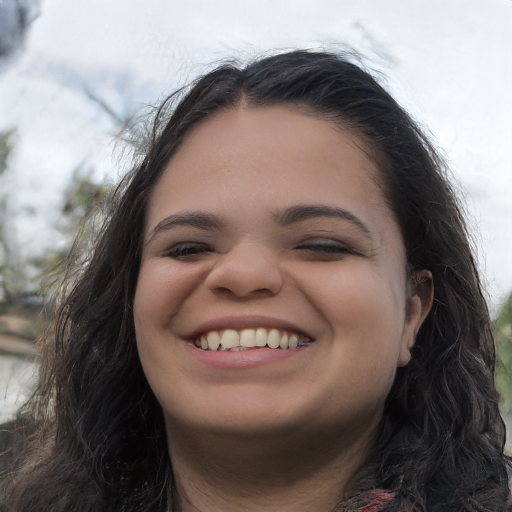}}
    % \hfil
    \subfloat[FreeDrag]{  \includegraphics[width=0.19\textwidth]{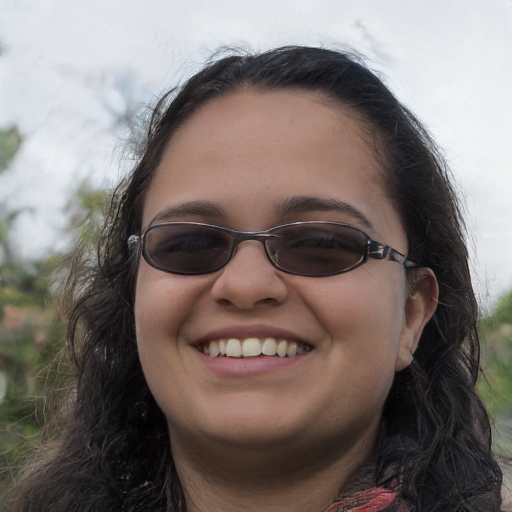}}
    % \hfil
    \subfloat[DragDiffusion]{  \includegraphics[width=0.19\textwidth]{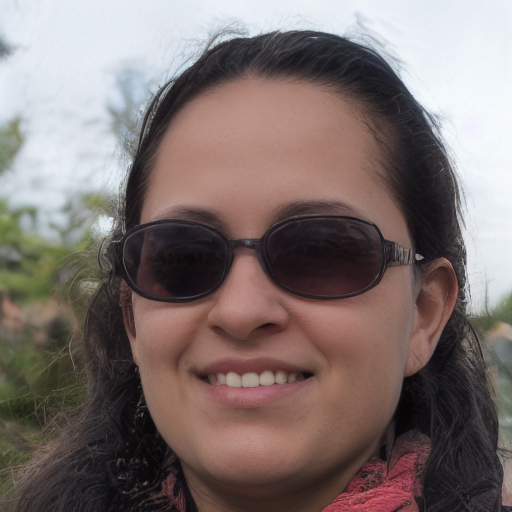}}
    % \label{fig:vsgan-face}
    \quad
    \subfloat{\includegraphics[width=0.19\textwidth]{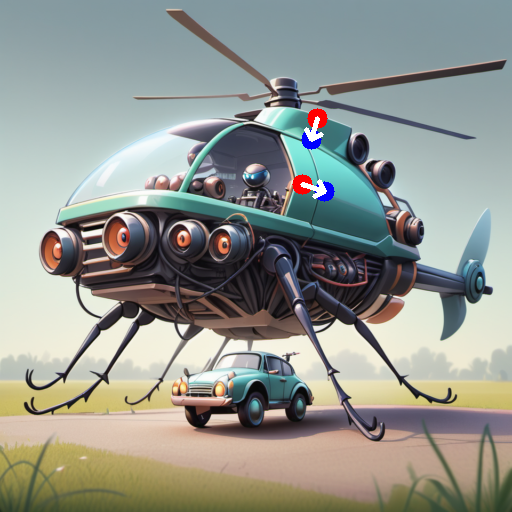}}
    % \hfill
    \hspace{2mm}
    \subfloat{  \includegraphics[width=0.19\textwidth]{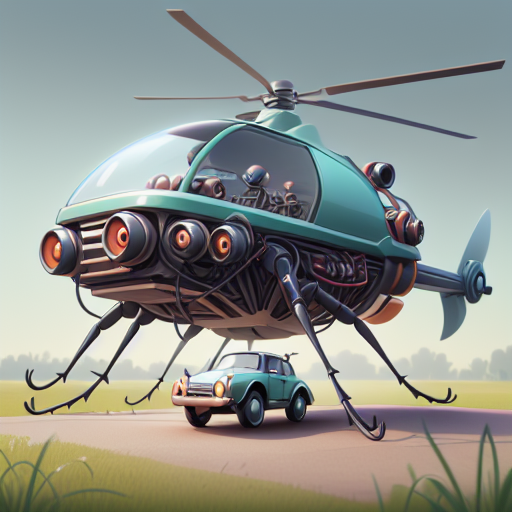}}
    % \hfill
    \subfloat{  \includegraphics[width=0.19\textwidth]{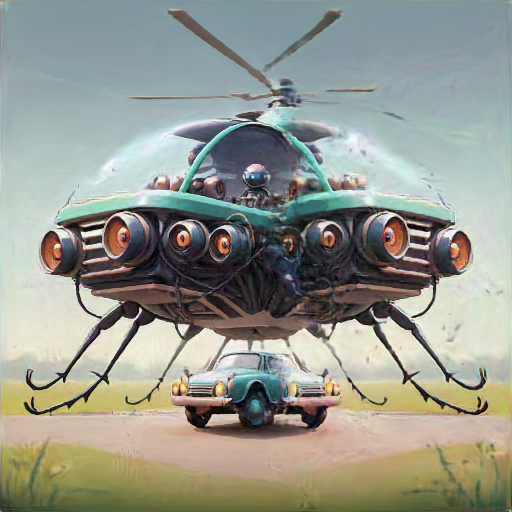}}
    % \hfill
    \subfloat{  \includegraphics[width=0.19\textwidth]{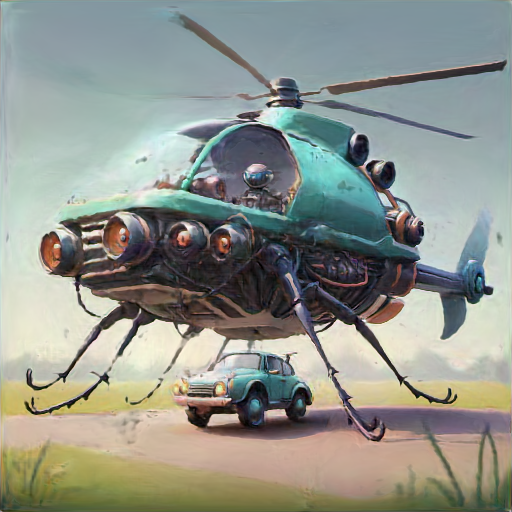}}
    % \hfill
    \subfloat{  \includegraphics[width=0.19\textwidth]{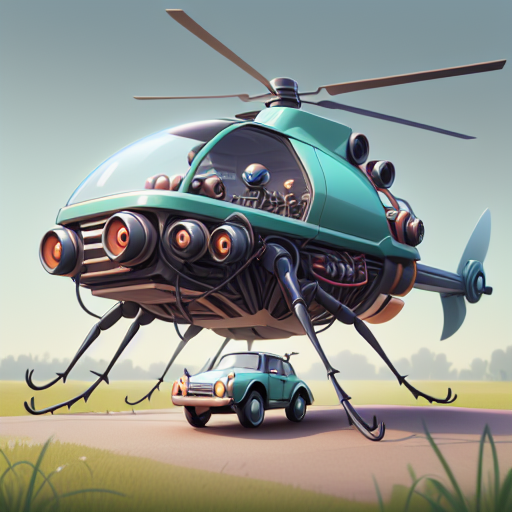}}
    % \label{fig:vsgan-break}
    \quad
    \subfloat{\includegraphics[width=0.19\textwidth]{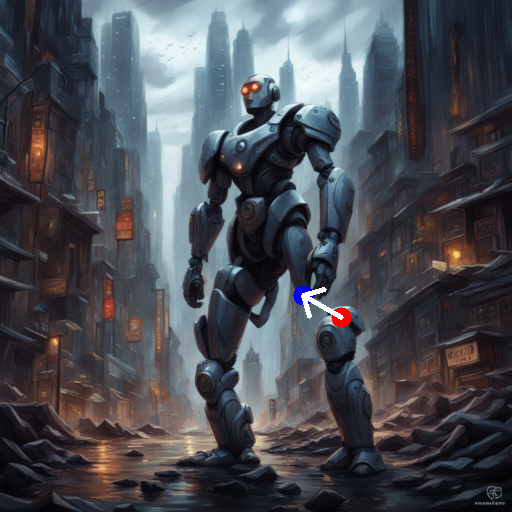}}
    % \hfill
    \hspace{4.3mm}
    \subfloat{  \includegraphics[width=0.19\textwidth]{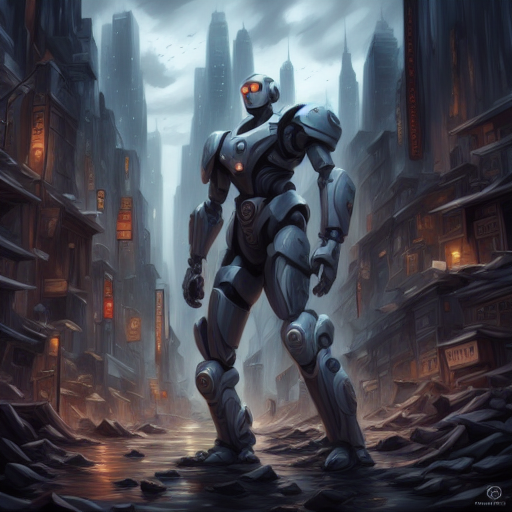}}
    % \hfill
    \subfloat{  \includegraphics[width=0.19\textwidth]{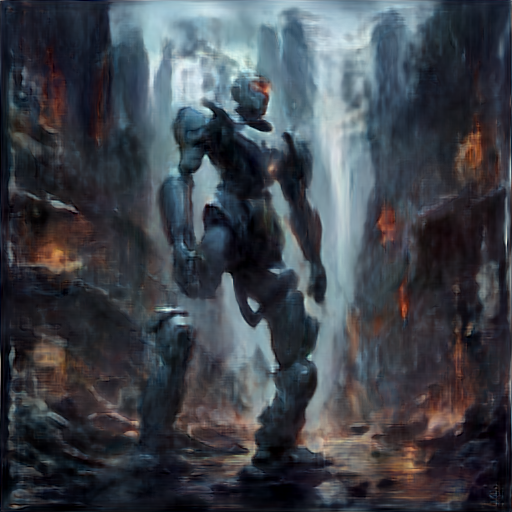}}
    % \hfill
    \subfloat{  \includegraphics[width=0.19\textwidth]{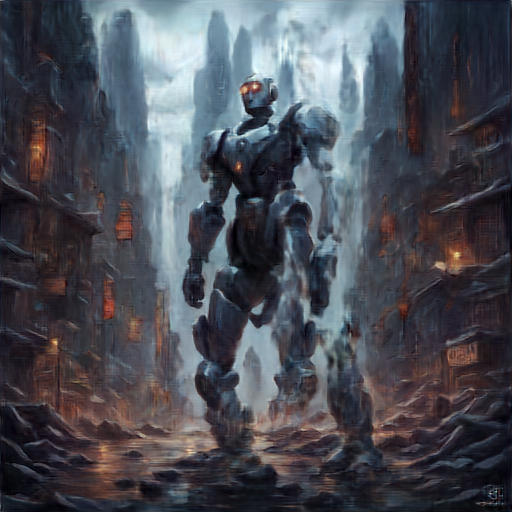}}
    % \hfill
    \subfloat{  \includegraphics[width=0.19\textwidth]{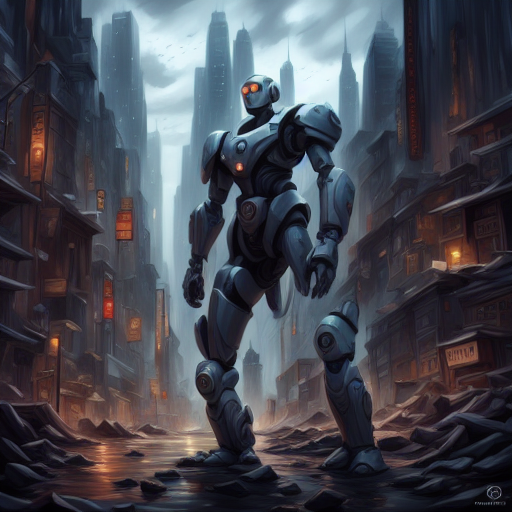}}
    % \label{fig:vsgan-model}
    \vspace{-2mm}\caption{Comparison of point-based editing methods with various drags. Our DragNoise exhibits superior semantic control ability.}\vspace{-5mm}
   \label{fig:vsgan}
 \end{figure*}

\subsubsection{Diffusion Semantic Propagation}
The manipulation noise $\hat \epsilon_\theta$ modifies the denoise direction for the image generation. However, we observe a forgetting issue where subsequent denoising processes tend to overlook the manipulation effect by simply performing diffusion semantic optimization on one timestep (see \cref{subsec:ablation}). As we have analyzed in \cref{fig:copy} that propagating the bottleneck feature to later timesteps does not have a significant influence on the overall semantics, we copy this optimized bottleneck feature $\hat s_t$ and substitute them in the subsequent timesteps. This is considered as the editing stage that keeps emphasizing the manipulation for the noise prediction. In the final few denoising timesteps, the structure of the image has essentially taken shape. Therefore, we stop replacing the bottleneck feature after timestep $t'$, where we treat it as the refinement stage.
After the diffusion semantic propagation in the denoising process, we obtain the generated output latent map $z'_0$ to produce the final image $x'_0$.

\subsection{Implementation Details}
We implement our method based on Stable Diffusion 1.5. Given an input image, to improve the image reconstruction, similar to~\cite{shi2023dragdiffusion}, we train a LoRA~\cite{hu2021lora} with the user-supplied image before diffusion semantic optimization. The fine-tuning process consists of 200 steps to update the LoRA's parameters.
{Given the need for users to edit anchor points on final generated images and for efficient editing, we treat all images, including diffusion-generated images, as ``real images'' with classifier-free guidance disabled. }We apply DDIM inversion to the given input image to attain the noisy latent map $z_t$ with $t=35$.
With the scheduled total timesteps of 50, during the diffusion semantic optimization stage, the beginning of the editing stage is therefore at $t=35$ and the refinement stage starts at $t'=10$. We employ the Adam optimizer with a learning rate of 0.01 for the semantic optimization. The default maximum number of optimization steps is 80. However, it is optional to increase the optimization steps in cases where the user encounters an exceptionally long dragging distance.
In \cref{eq:loss_align} and \cref{eq:tracking}, we set $r_1$ to 1 and $r_2$ to 3. The value of $\lambda$ in \cref{eq:objective} is set to 0.1.

\begin{figure*}

    \captionsetup[subfloat]{position=top}
    \subfloat[Input]{\includegraphics[width=0.16\textwidth]{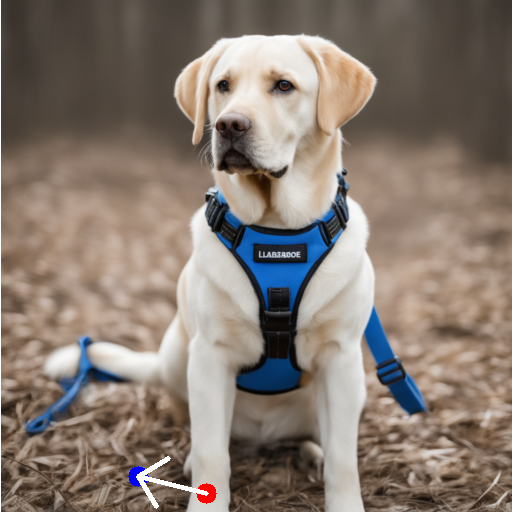}}
    \subfloat[DragNoise]{\label{fig:vsdiff-dragnoise1}  \includegraphics[width=0.16\textwidth]{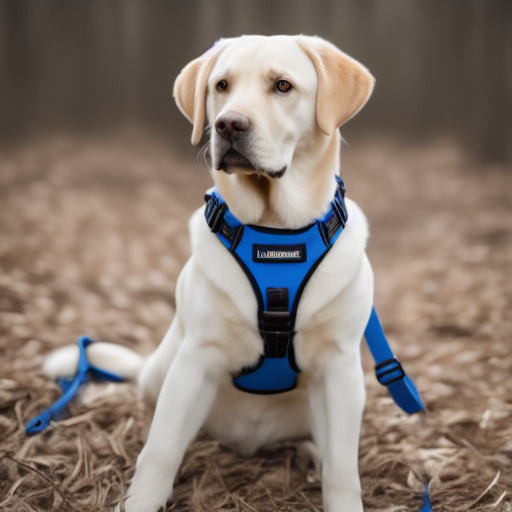}}
    \subfloat[DragDiffusion]{\label{fig:vsdiff-dragdiff1}  \includegraphics[width=0.16\textwidth]{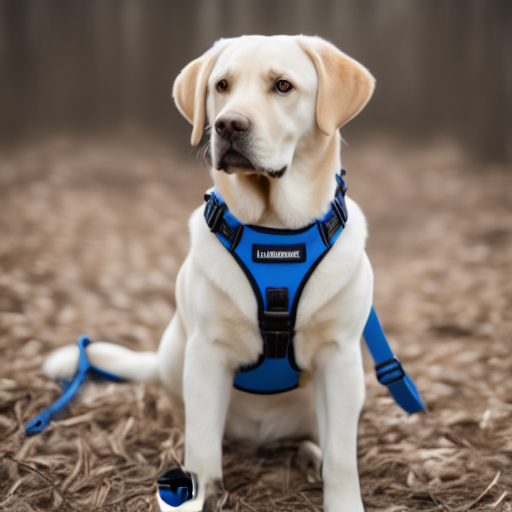}}
    \subfloat[Input]{  \includegraphics[width=0.16\textwidth]{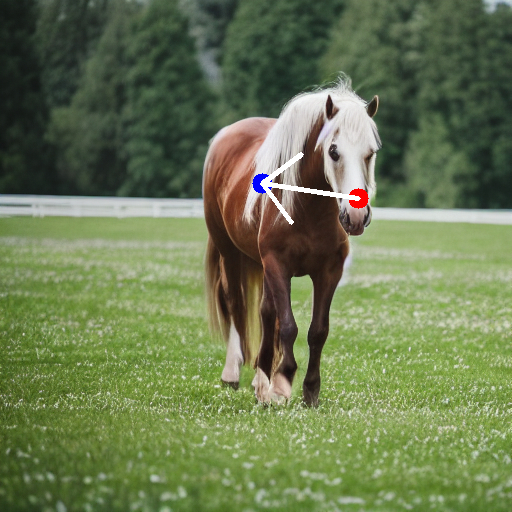}}
    \subfloat[DragNoise]{  \includegraphics[width=0.16\textwidth]{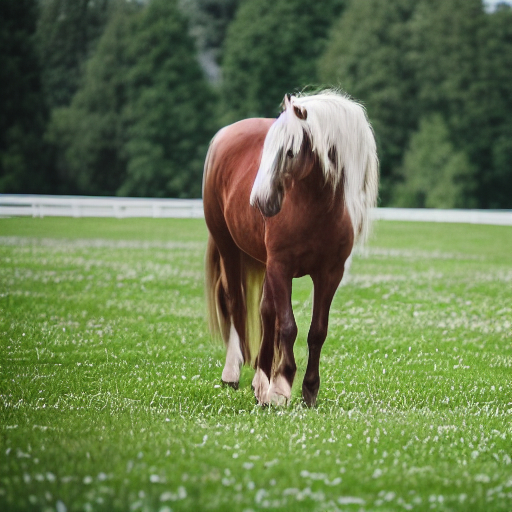}}
    \subfloat[DragDiffusion]{\label{fig:vsdiff-dragdiff2} \includegraphics[width=0.16\textwidth]{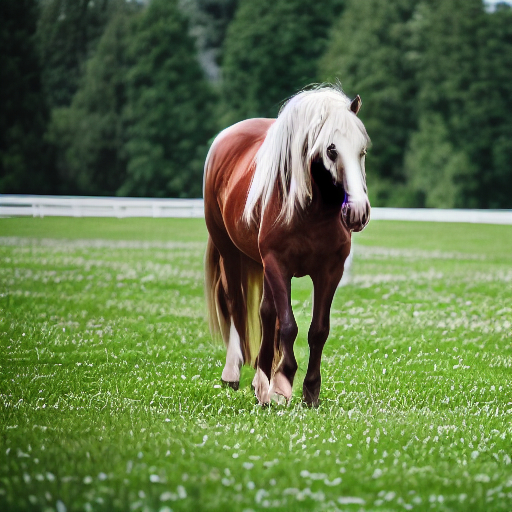}}
    % \label{fig:vsgan-cat}
    \quad

    \subfloat{\includegraphics[width=0.16\textwidth]{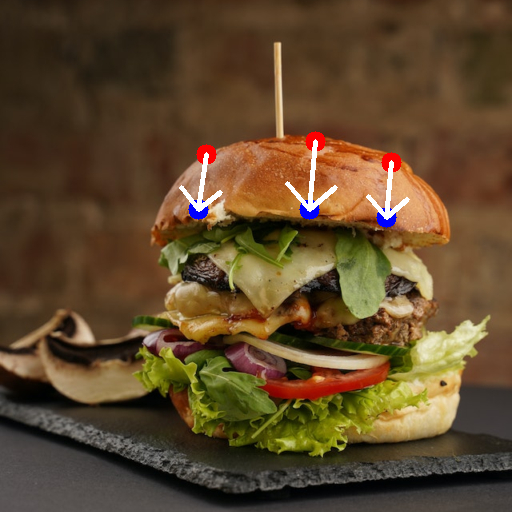}}
    \subfloat{  \includegraphics[width=0.16\textwidth]{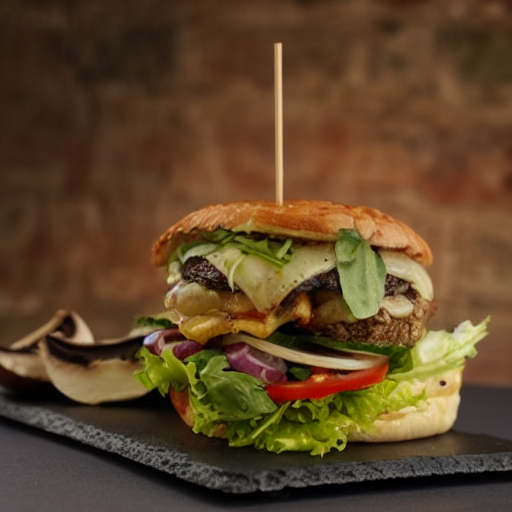}}
    \subfloat{  \includegraphics[width=0.16\textwidth]{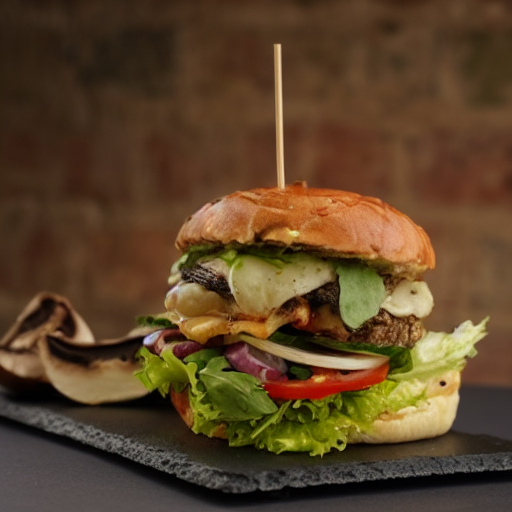}}
    \subfloat{  \includegraphics[width=0.16\textwidth]{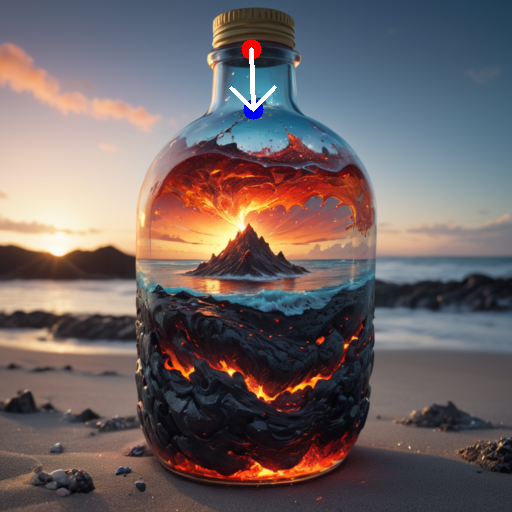}}
    \subfloat{  \includegraphics[width=0.16\textwidth]{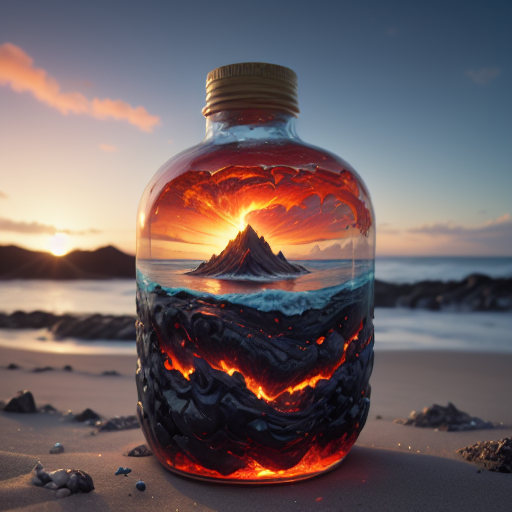}}
    \subfloat{  \includegraphics[width=0.16\textwidth]{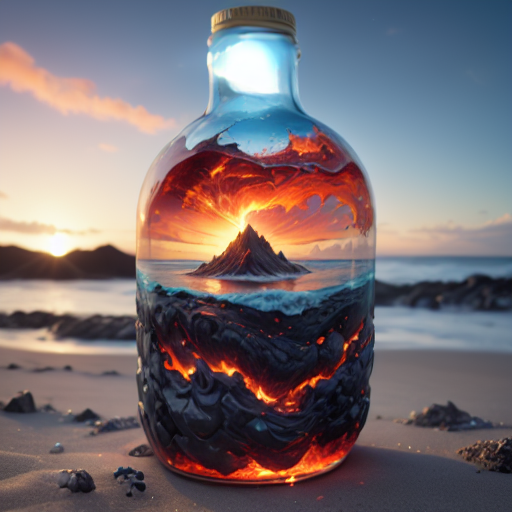}}
    % \label{fig:vsgan-cat}
    \quad

    \subfloat{\includegraphics[width=0.16\textwidth]{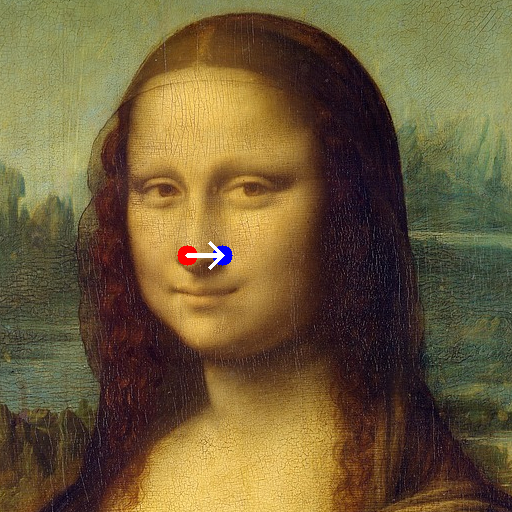}}
    \subfloat{  \includegraphics[width=0.16\textwidth]{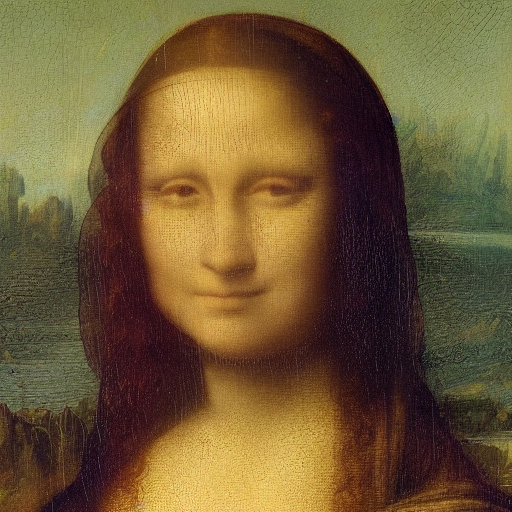}}
    \subfloat{  \includegraphics[width=0.16\textwidth]{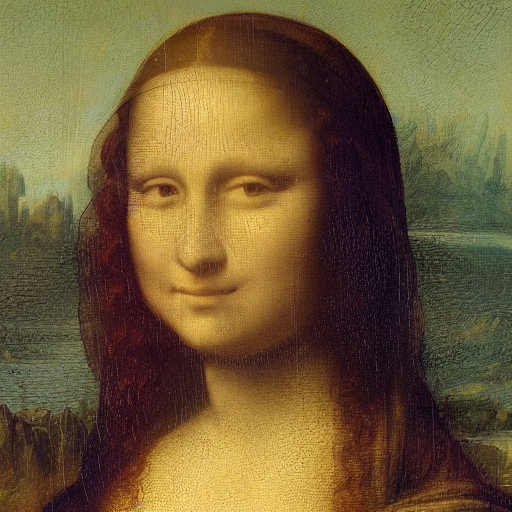}}
    \subfloat{  \includegraphics[width=0.16\textwidth]{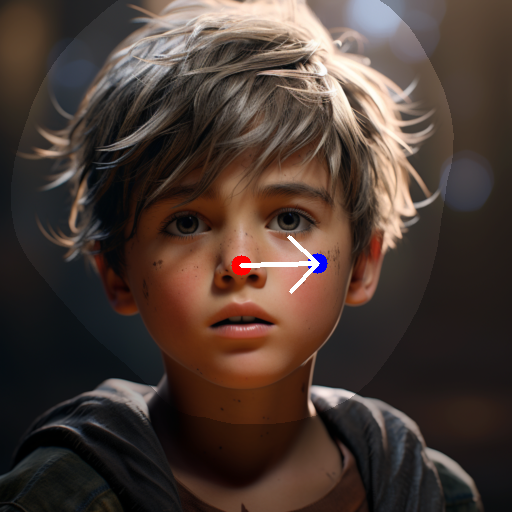}}
    \subfloat{  \includegraphics[width=0.16\textwidth]{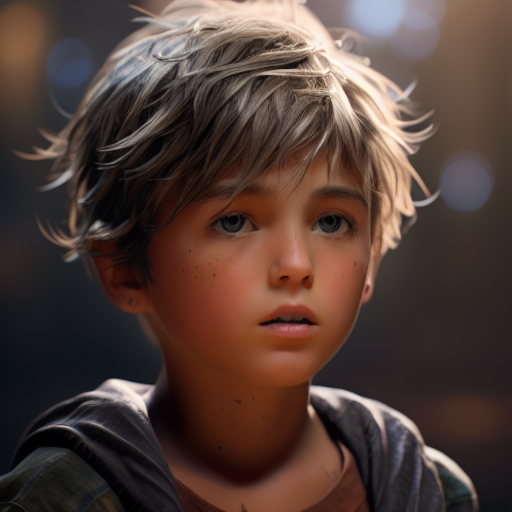}}
    \subfloat{  \includegraphics[width=0.16\textwidth]{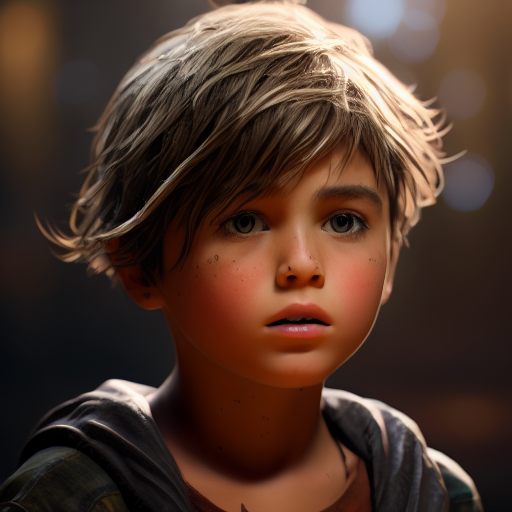}}
    % \label{fig:vsgan-cat}
    \quad

    \subfloat{\includegraphics[width=0.16\textwidth]{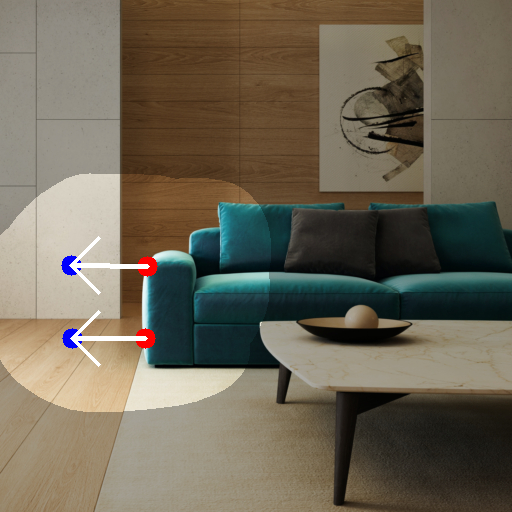}}
    \subfloat{  \includegraphics[width=0.16\textwidth]{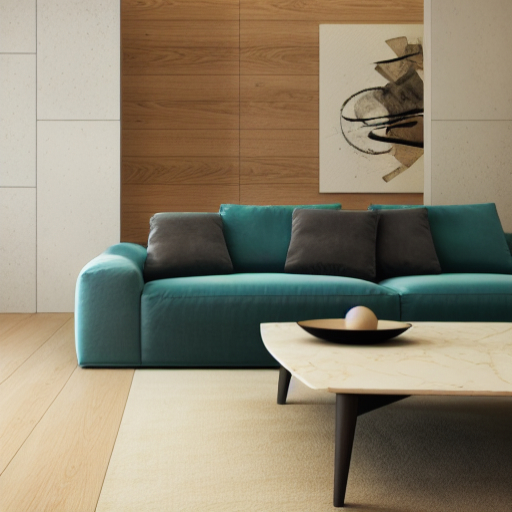}}
    \subfloat{  \includegraphics[width=0.16\textwidth]{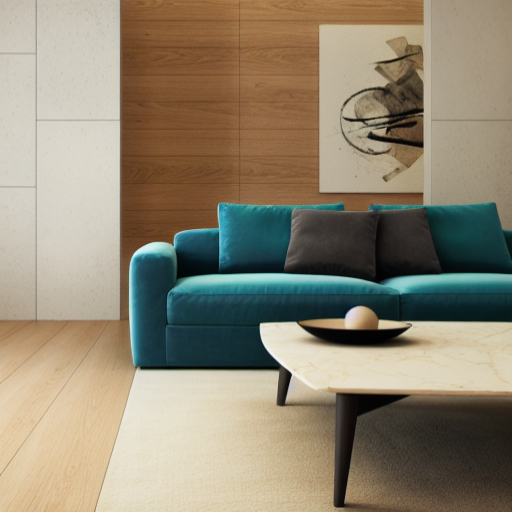}}
    \subfloat{  \includegraphics[width=0.16\textwidth]{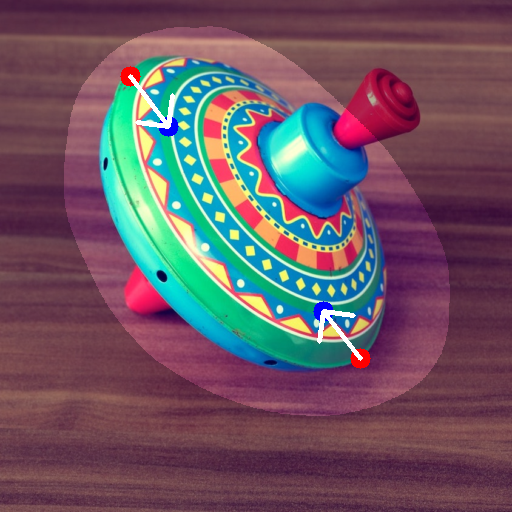}}
    \subfloat{  \includegraphics[width=0.16\textwidth]{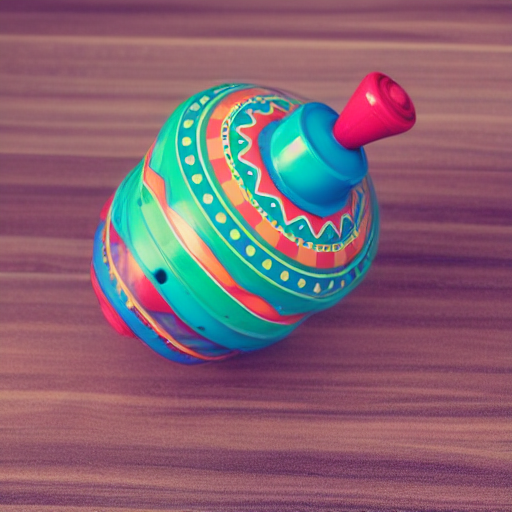}}
    \subfloat{  \includegraphics[width=0.16\textwidth]{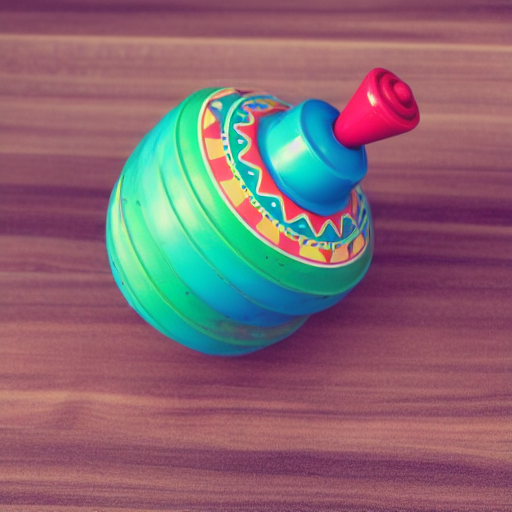}}
    % \label{fig:vsgan-cat}
    \quad

    \vspace{-2mm}\caption{Comparison with DragDiffusion across diverse images. Our method yields more stable and plausible edits aligned with user inputs.}
    \label{fig:vsDiffusion}
    \vspace{-5mm}
 \end{figure*}

\section{Experiments}
\label{sec:experiments}
In this section, we conduct experiments on the drag-based editing benchmark, DragBench~\cite{shi2023dragdiffusion}, as well as diverse example images to assess the effectiveness of DragNoise.
We first compare DragNoise with existing GAN-based and diffusion-based methods. Furthermore, we validate the optimization efficiency of our method by comparing it with recent DragDiffusion. Additionally, ablation studies are conducted to analyze the influences of different initial timesteps of the editing stage, the effect of optimization on distinct layers, and the extent of optimization propagation at the editing stage. Quantitative experiments using various metrics are carried out to validate our dragging accuracy and image fidelity. %{Lastly, we showcase multi-point controlled examples to demonstrate DragNoise's adaptability across various editing scenarios.}

\subsection{Qualitative Evaluations}

In this subsection, we compare our method to existing approaches, \ie, DragGAN~\cite{pan2023drag}, FreeDrag~\cite{ling2023freedrag} and DragDiffusion~\cite{shi2023dragdiffusion}. For all the input images, we train StyleGAN and LoRA for GAN-based editing and diffusion-based editing, respectively.

In \cref{fig:vsgan}, we conduct a general comparison with GAN-based methods and the diffusion-based method.
Both DragGAN and FreeDrag exhibit the capability to relocate anchor points to objective points. However, due to the limited spatial information in their 1D latent space, they usually encounter global changes during the dragging process. For instance, the movement of anchor points results in the unintended shifting of features such as the person's facial shape in the 1st row.
{Moreover, DragGAN incorrectly tracks new anchor points when the features around the anchor points are similar, leading to unexpected editing effects such as flipping and mirroring (see \cref{fig:vsgan-draggan}).}
The results show that both GAN-based methods are constrained by the limitations in StyleGAN's generation and inversion capabilities. As a consequence, both DragGAN and FreeDrag yield outcomes where the regions around anchor points become blurred when additional contents are required to be generated to fill the space.

We further showcase diverse results in comparison with DragDiffusion in \cref{fig:vsDiffusion}. Results show that DragDiffusion's two-dimensional noisy latent map offers improved local editing capabilities without inducing global changes. Nevertheless, the inversion capability of the noisy latent map is constrained, resulting in a loss of semantic information. Moreover, the lack of highly decoupled semantic information in the noisy latent map results in semantic alterations in the generated images around the anchor points. For example, as demonstrated in the 3rd row of \cref{fig:vsdiff-dragdiff2}, the hairstyle changes after the face rotation. In addition, due to the long chain of back-propagation of DragDiffusion's optimization on the noisy latent map, if the features near the anchor points are similar, \ie, when the semantic alignment loss is small, the gradient vanishes, impeding updates to the noisy latent map and consequently leading to an under-dragged image (\eg, the horse in \cref{fig:vsdiff-dragdiff2}).

In comparison, our method precisely moves the anchor points toward the objective points and achieves a natural dragging effect with high fidelity. This is attributed to the proposed diffusion semantic optimization and propagation with the bottleneck feature. For instance, when dragging to narrow the sunglasses, our modifications do not change the face shape (see the 1st row of \cref{fig:vsgan-dragnoise}). In both \cref{fig:vsgan,fig:vsDiffusion}, our method exhibits superior editing capabilities, ensuring more accurate dragging even amidst substantial changes while also preserving semantics or identities more effectively. Generally, when moving an object, filling a space or resizing an object, changes are confined to the contents associated with the anchor points, preserving the unchanged status of the other semantic contents.

\begin{figure}[t]
    \centering
    \begin{subfigure}{.24\linewidth}
    \centering
      \includegraphics[width=1.0\textwidth]{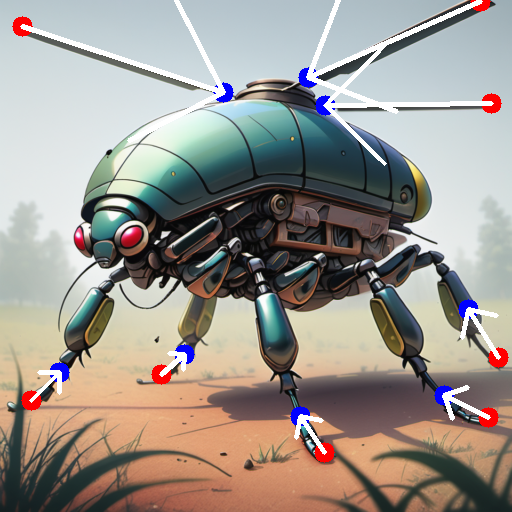}
    \caption{Input}
    \end{subfigure}
    % \hspace{-1.5mm}
    \begin{subfigure}{.24\linewidth}
        \centering
        \includegraphics[width=1.0\textwidth]{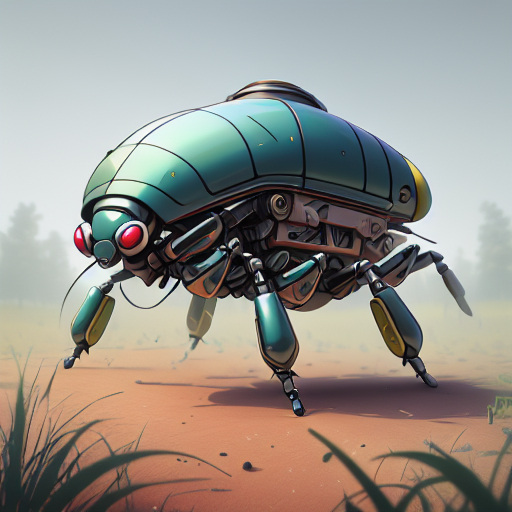}
    \caption{Result}
    \label{multipoint-b}
    \end{subfigure}
    % \hspace{-1.5mm}
    \begin{subfigure}{.24\linewidth}
        \centering
        \includegraphics[width=1.0\textwidth]{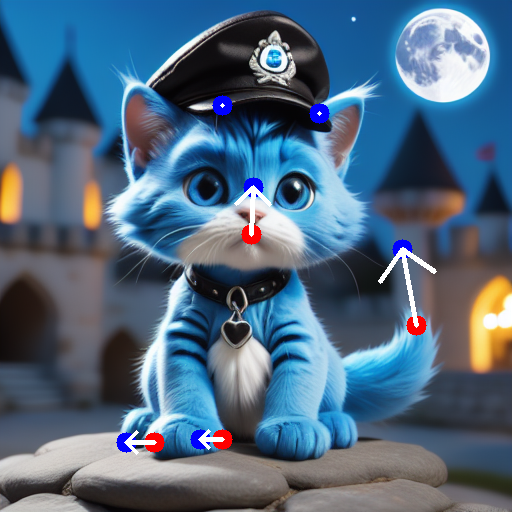}
    \caption{Input}
    \end{subfigure}
    % \hspace{-1.5mm}
    \begin{subfigure}{.24\linewidth}
        \centering
        \includegraphics[width=1.0\textwidth]{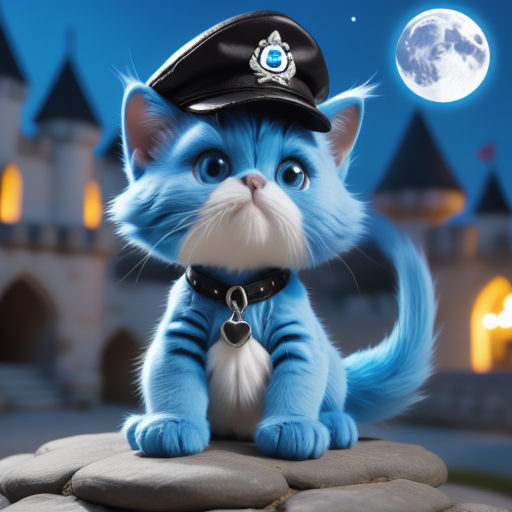}
        \caption{Result}
        \label{multipoint-d}
    \end{subfigure}
    \vspace{-3mm}\caption{Our DragNoise also supports multi-point editing.}\vspace{-5mm}
    \label{fig:multi-point}
\end{figure}

{In \cref{fig:multi-point}, we further showcase multi-point control with two examples: precise editing of propellers and legs (\cref{multipoint-b}), and lifting a cat's head while keeping its hat in place (\cref{multipoint-d}). These cases with multi-points for multi-targets highlight DragNoise's versatility in diverse editing scenarios.}

\subsection{Analysis of Optimization Efficiency}
\begin{figure}[t]
   \subfloat[Input]{\label{speed-a}\includegraphics[width=0.15\textwidth]{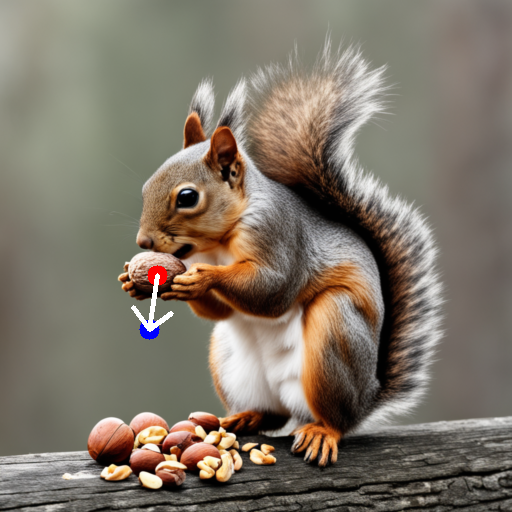}}
   \hfil
   \subfloat[DragNoise]{\label{speed-b}\includegraphics[width=0.15\textwidth]{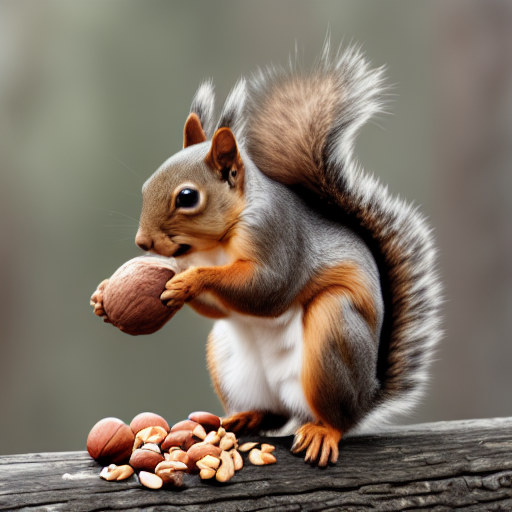}}
   \hfil
   \subfloat[DragDiffusion]{\label{speed-c}\includegraphics[width=0.15\textwidth]{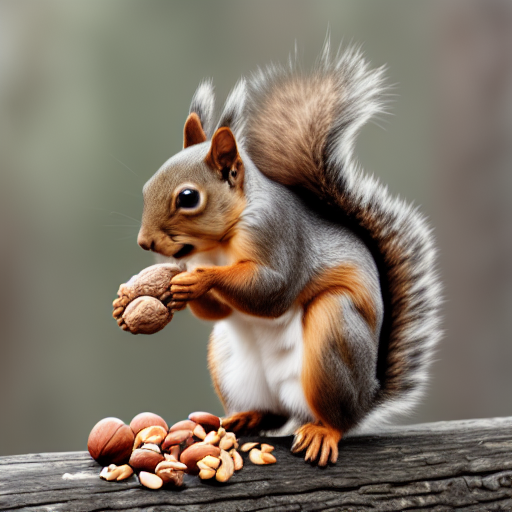}}
   \quad
   \subfloat[Optimization Loss]{\label{speed-d}\includegraphics[width=0.23\textwidth]{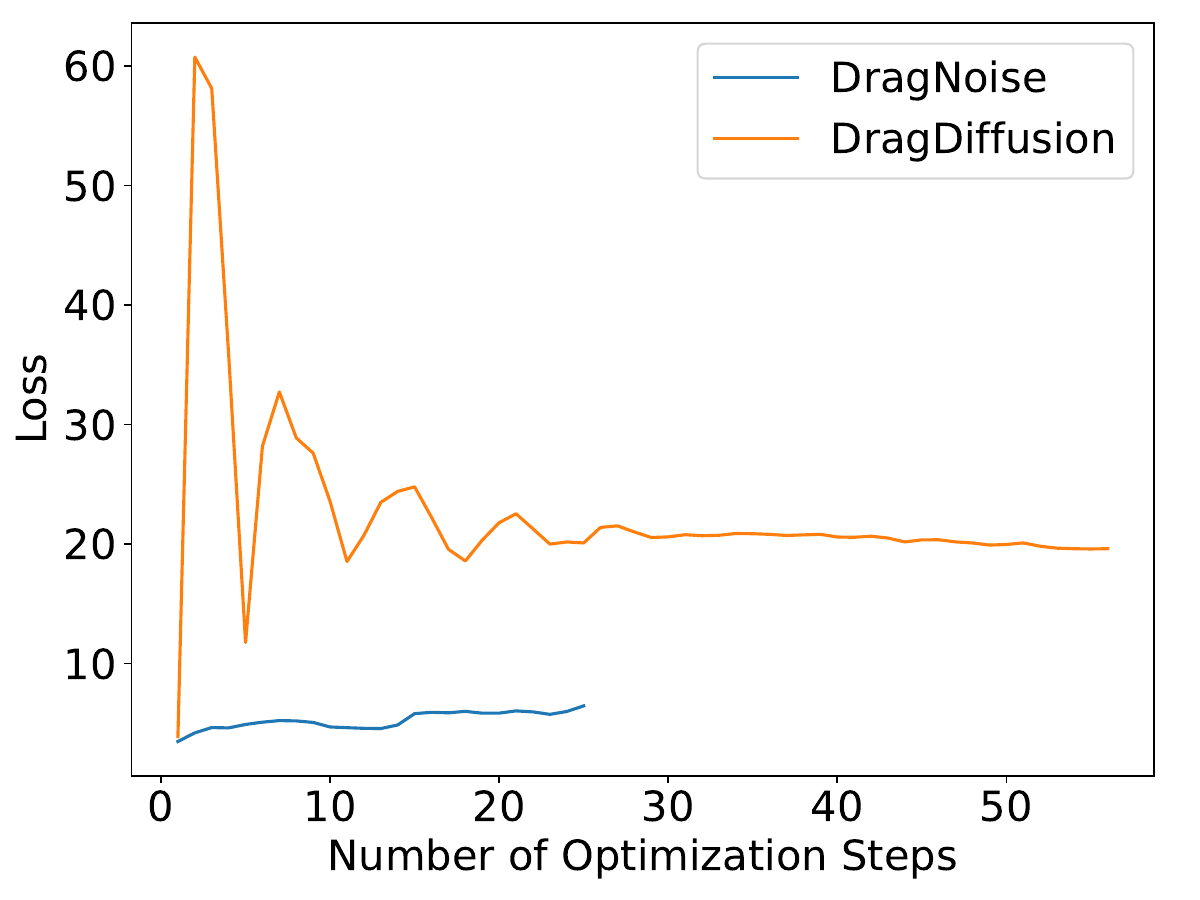}}
   \subfloat[Motion Trajectory of Anchor Points]{\label{speed-e}\includegraphics[width=0.26\textwidth]{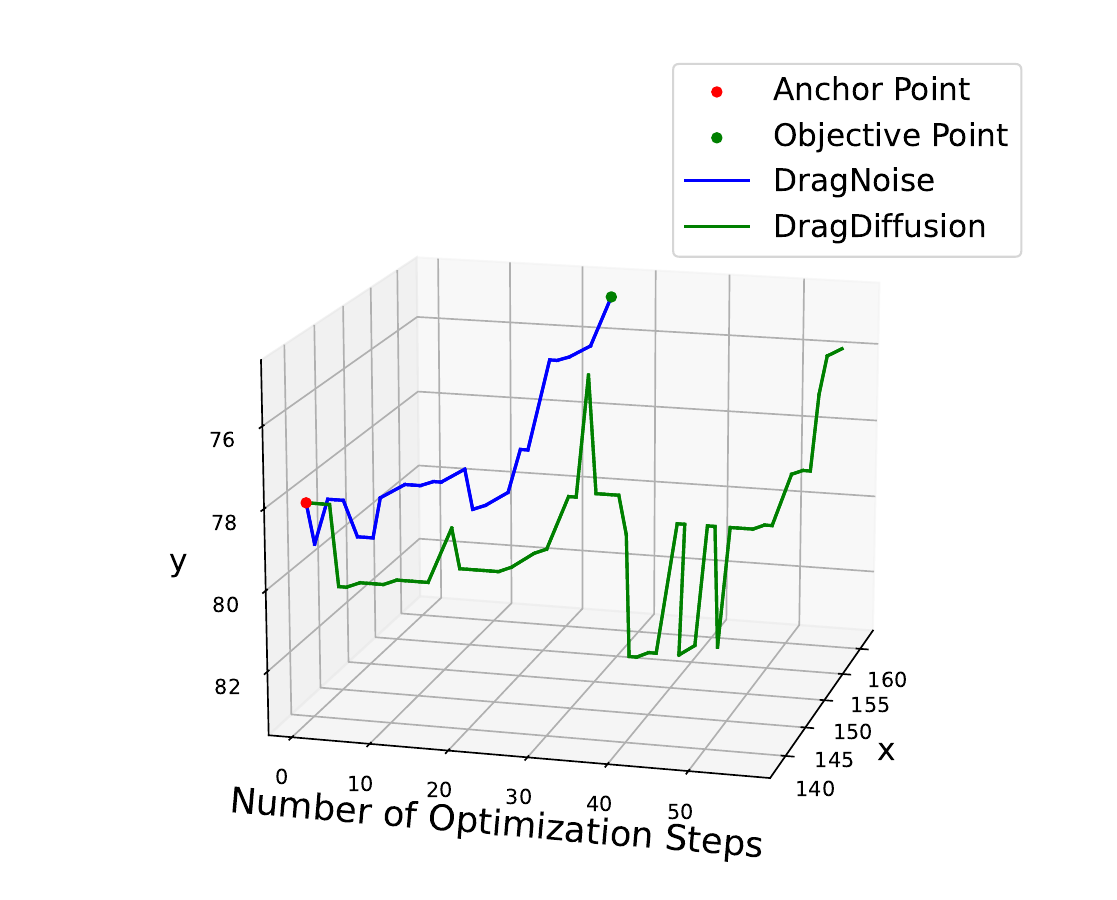}}
   \vspace{-3mm}
   \caption{Comparison results of optimization efficiency. (a) and (b) are edited results of the DragNoise and DragDiffusion. (d)) and (e) illustrate the change in total loss and the motion trajectory of the anchor points during the optimization.}
   \vspace{-5mm}
   \label{fig:speed}
\end{figure}

\cref{fig:speed} demonstrates an example of the optimization process of DragNoise and DragDiffusion, depicting their respective changes in loss and the trajectory of the anchor point. DragNoise rapidly determines the direction of semantic editing in the bottleneck feature, approaching the objective point with the optimization process completed in merely 25 steps. In contrast, DragDiffusion requires 56 iterations to reach the objective point.

The optimized bottleneck feature in DragNoise offers a more enriched and higher-level semantic space in contrast to the latent map utilized in DragDiffusion. Consequently, the direction of semantic editing is discernible at the start, leading to a reduction in the number of optimization steps by over 50\%. In addition, within a single optimization iteration, DragNoise achieves a 10\% reduction in the time required compared to DragDiffusion, attributed to its shorter back-propagation chain. {This results in DragNoise cutting down the optimization time by over 50\%.} On the Tesla V100 graphics card, editing an image with a resolution of $512 \times 512$ takes approximately 10 seconds for DragNoise, whereas DragDiffusion requires over 22 seconds.

\subsection{Ablation Study}
\label{subsec:ablation}
\begin{figure*}
    \subfloat[Input]{\includegraphics[width=0.19\textwidth]{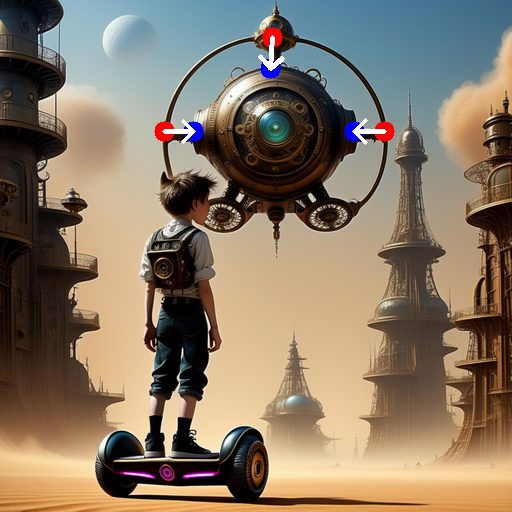}}
    \hfil
    \subfloat[t=45]{\includegraphics[width=0.19\textwidth]{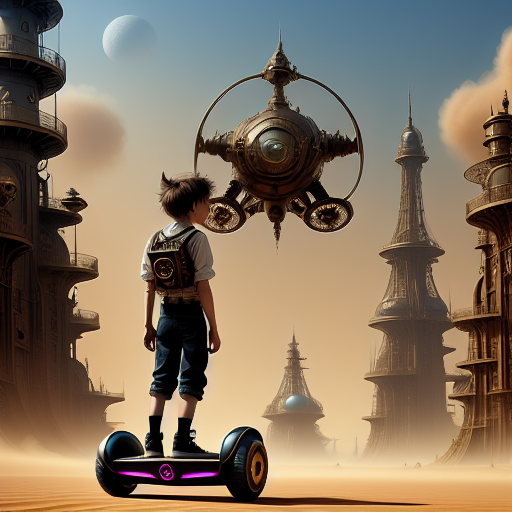}}
    \hfil
    \subfloat[t=40]{\includegraphics[width=0.19\textwidth]{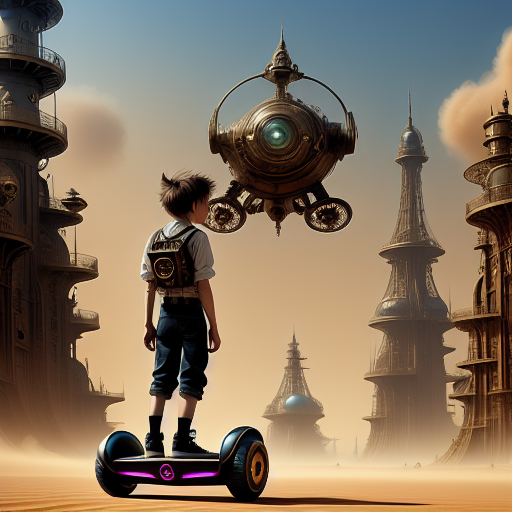}}
    \hfil
    \subfloat[t=35]{\includegraphics[width=0.19\textwidth]{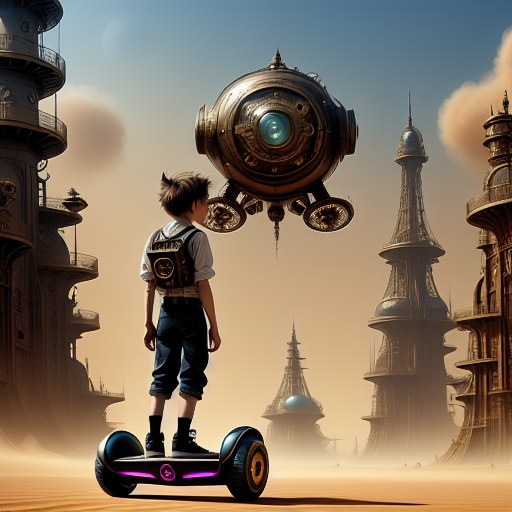}}
    \hfil
    \subfloat[t=30]{\includegraphics[width=0.19\textwidth]{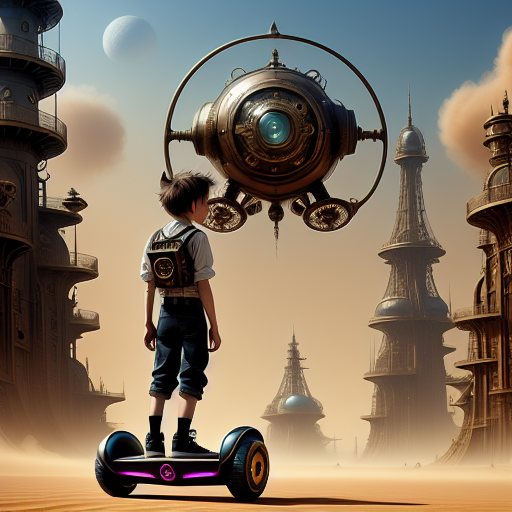}}
    \vspace{-4mm}\caption{Results of performing the same dragging operation while initiating the editing process with different initial timesteps.}\vspace{-3mm}
    \label{fig:inversion}
 \end{figure*}

\textbf{Effect of the initial timesteps of the editing stage.}
We study the effect of optimizing the bottleneck feature at different beginnings of the editing stage. \cref{fig:inversion} shows the effect of the same dragging operation with different initial editing timesteps. The results show that with earlier initial timesteps, \eg, $t=45$, the bottleneck feature controls the larger semantics, shrinking the whole object. With later initial editing timesteps, the smaller the semantics can be controlled, \eg, at $t=35$ the outer ring of iron was reduced. However, at $t=30$ only the iron around the anchor points are reduced. This experiment shows the flexibility of our method to control the semantics of various scales.

\begin{figure*}
    \subfloat[Input]{\includegraphics[width=0.16\textwidth]{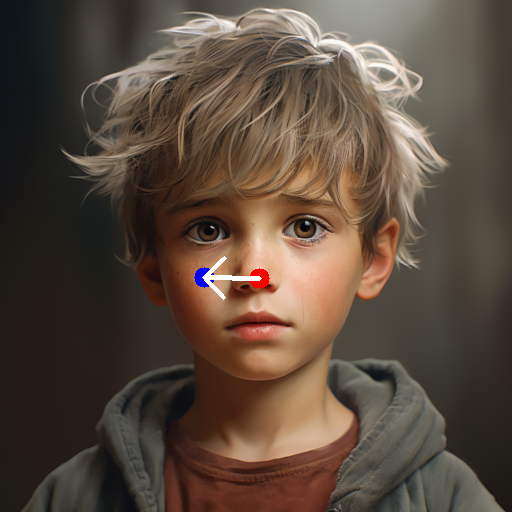}}
    \hfill
    \subfloat[Encoder Block 1]{\includegraphics[width=0.16\textwidth]{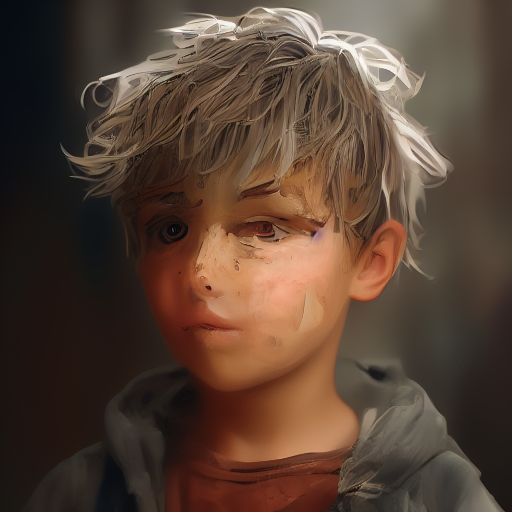}}
    \hfill
    \subfloat[Encoder Block 3]{\includegraphics[width=0.16\textwidth]{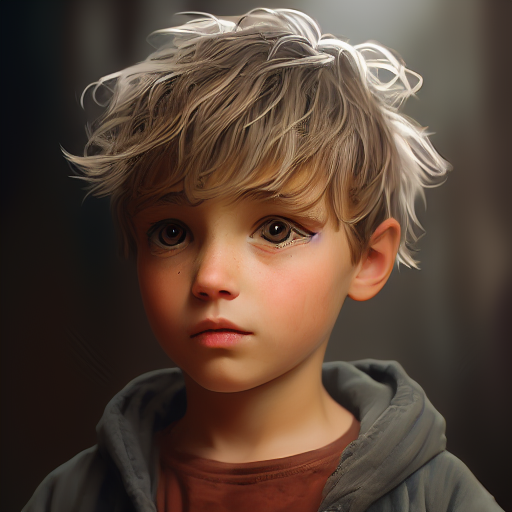}}
    \hfill
    \subfloat[Bottleneck]{\includegraphics[width=0.16\textwidth]{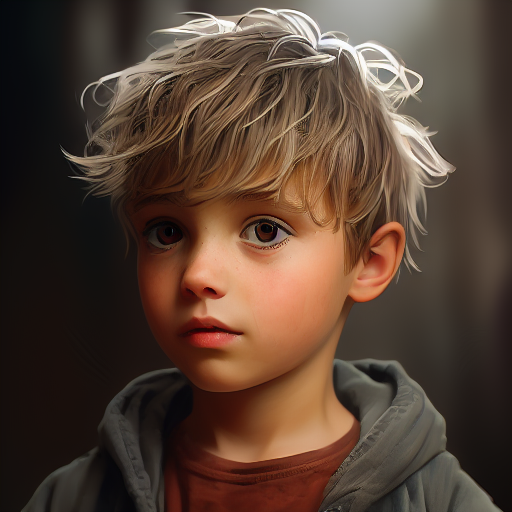}}
    \hfill
    \subfloat[Decoder Block 1]{\includegraphics[width=0.16\textwidth]{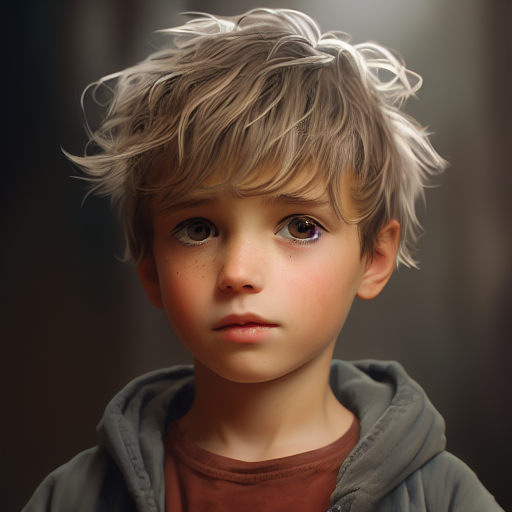}}
    \hfill
    \subfloat[Decoder Block 3]{\includegraphics[width=0.16\textwidth]{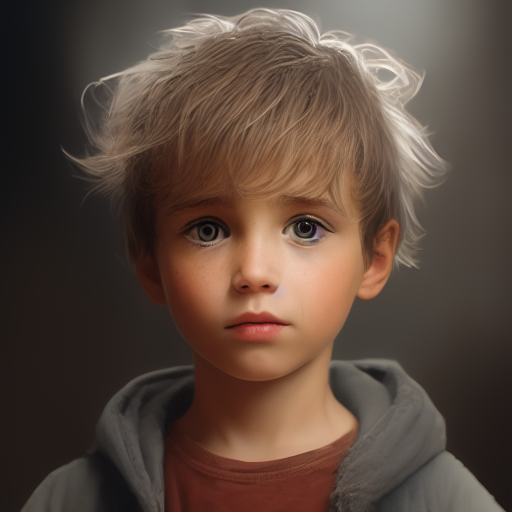}}
    \vspace{-3mm}\caption{Results of different feature layers in U-Net using as optimization targets.}\vspace{-5mm}
    \label{fig:updatelayer}
 \end{figure*} 
\textbf{Effect of optimizing different features.}
We analyze the impact of updating different features from the U-Net when dragging the image while keeping other settings as default. As shown in \cref{fig:updatelayer}, optimizing the bottleneck feature yields the optimal results in terms of image quality and controllability, allowing for the most precise control. Optimizing the encoder blocks leads to poor generation quality as the encoder features struggle to capture complete semantics. On the other hand, optimizing decoder blocks faces challenges in determining the direction of semantic editing.

\begin{figure}[h]
    \centering
    \begin{subfigure}{.24\linewidth}
    \centering
      \includegraphics[width=1.0\textwidth]{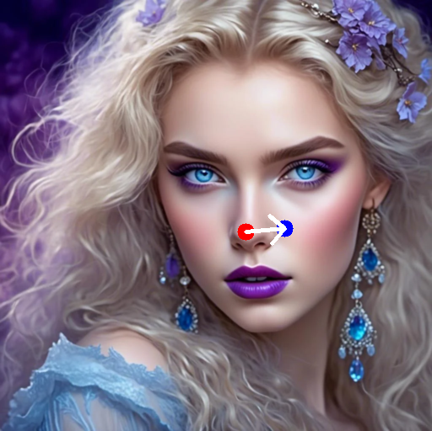}
    \caption{Input}
    \label{endtime-a}
    \end{subfigure}
    % \hspace{-1.5mm}
    \begin{subfigure}{.24\linewidth}
        \centering
        \includegraphics[width=1.0\textwidth]{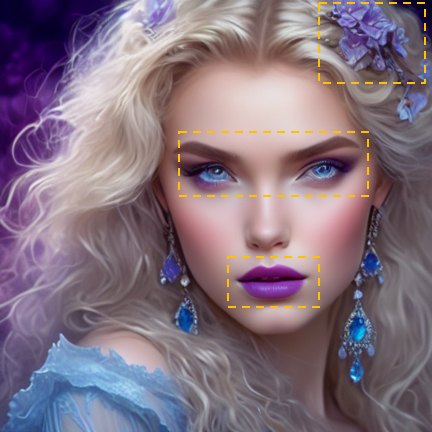}
    \caption{Halt at t'=20}
    \label{endtime-b}
    \end{subfigure}
    % \hspace{-1.5mm}
    \begin{subfigure}{.24\linewidth}
        \centering
        \includegraphics[width=1.0\textwidth]{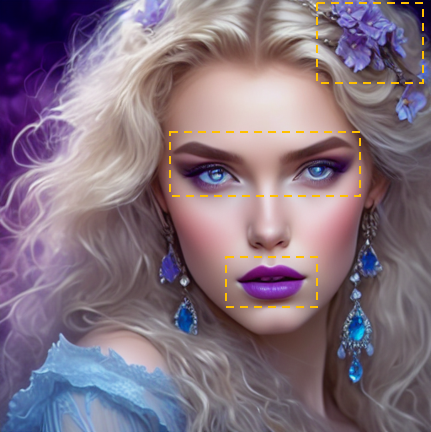}
    \caption{Halt at t'=10}
    \label{endtime-c}
    \end{subfigure}
    % \hspace{-1.5mm}
    \begin{subfigure}{.24\linewidth}
        \centering
        \includegraphics[width=1.0\textwidth]{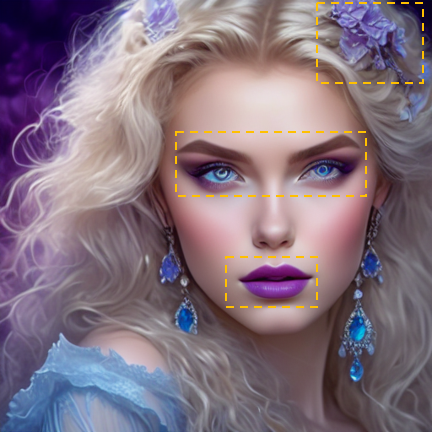}
        \caption{No Halting}
        \label{endtime-d}
    \end{subfigure}
    \vspace{-3mm}\caption{Copying optimized bottleneck features into the denoising process and halting propagation at different timesteps.}\vspace{-5mm}
    \label{fig:copy-endtime}
\end{figure}

\textbf{Effect of different extents of the editing stage.}
We analyze the effect of different extents of the editing stage with other settings as default. {In \cref{fig:copy-endtime}, selecting a proper editing endpoint is crucial for image generation quality. Editing until $t'=20$ leads to instability and vagueness, whereas controlling noise until $t'=10$ ensures clarity and detail preservation. Beyond $t'=10$, the last denoising steps reintroduce fine original image details without the optimized bottleneck feature's influence. Conversely, controlling predicted noise throughout can result in the loss of details like teeth or flower textures, yielding unclear outputs.}

\subsection{Quantitative Analysis}
\begin{figure}
    \centering
    \subfloat{\includegraphics[width=0.4\textwidth]{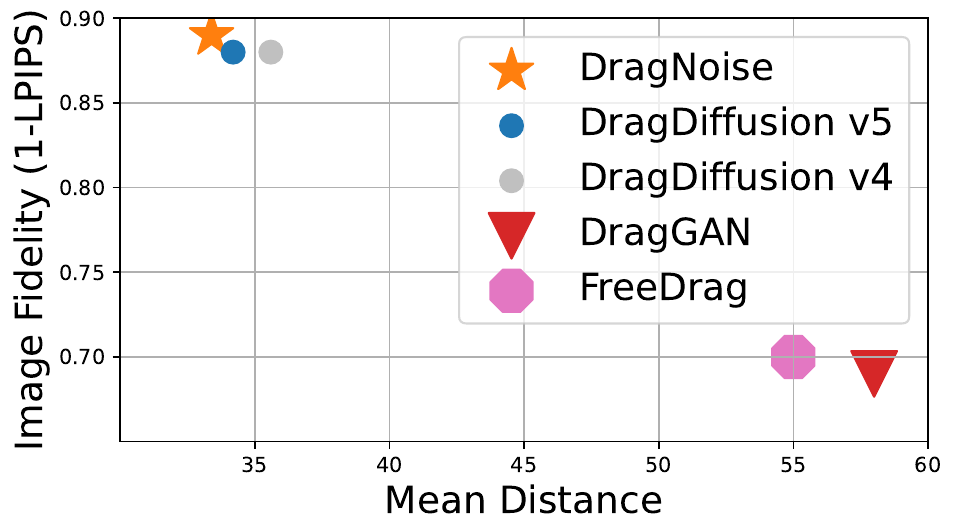}}
    \vspace{-3mm}\caption{Quantitative evaluations on DragBench, we achieve the best editing accuracy and fidelity.}\vspace{-3mm}
    \label{fig:Quantative}
 \end{figure}

We adopt the DragBench dataset and quantitative metrics of mean distance (MD)~\cite{tang2023dift,pan2023drag,shi2023dragdiffusion} and image fidelity (IF)~\cite{kawar2023imagic} for quantitative analyses. IF serves as an indicator of the fidelity of editing outcomes, and MD reflects the accuracy of the dragging effect. \cref{fig:Quantative} demonstrates the performance of the four methods on the DragBench dataset, the diffusion-based editing methods generally outperform the GAN-based methods. Notably, our DragNoise, surpasses all existing methods in terms of dragging accuracy and image fidelity.

\section{Conclusion, Limitation, and Future Work}
\label{sec:conclusion}
In this paper, we propose DragNoise, a novel interactive point-based image editing method that leverages diffusion semantic propagation. Drawing upon a thorough exploration of diffusion semantics, we consider the predicted noise in the reverse diffusion process as semantic editors. Our approach involves diffusion semantic optimization and diffusion semantic propagation, enabling the editing of diffusion semantics within a single timestep while effectively propagating the resultant changes. Extensive experiments have validated our method's efficiency and flexibility.

\begin{figure}[t]
    \centering
    \begin{subfigure}{.24\linewidth}
    \centering
      \includegraphics[width=1.0\textwidth]{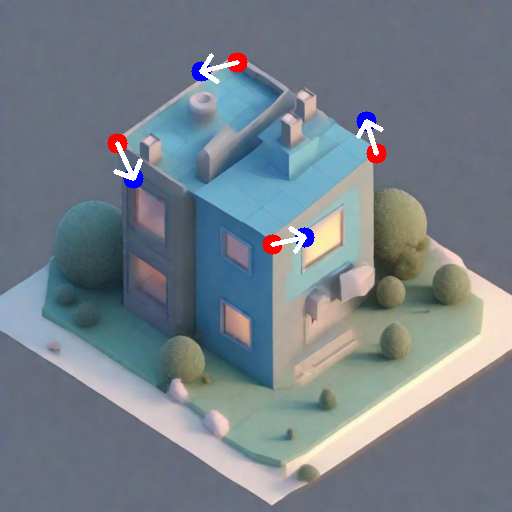}
    \caption{Input}
    \label{failure-a}
    \end{subfigure}
    % \hspace{-1.5mm}
    \begin{subfigure}{.24\linewidth}
        \centering
        \includegraphics[width=1.0\textwidth]{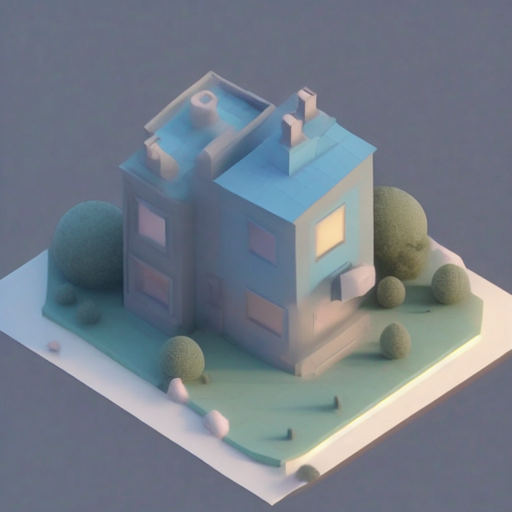}
    \caption{Result}
    \label{failure-b}
    \end{subfigure}
    % \hspace{-1.5mm}
    \begin{subfigure}{.24\linewidth}
        \centering
        \includegraphics[width=1.0\textwidth]{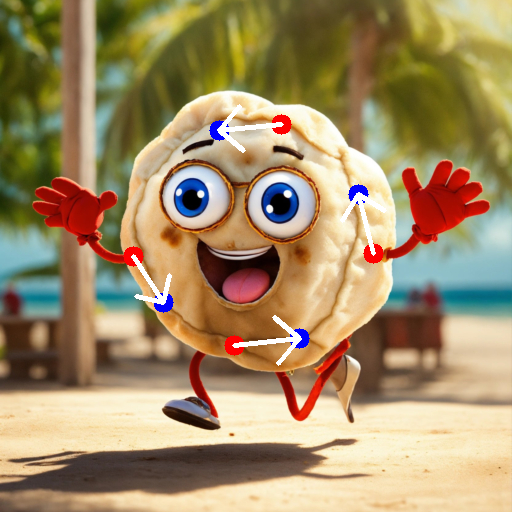}
    \caption{Input}
    \label{failure-c}
    \end{subfigure}
    % \hspace{-1.5mm}
    \begin{subfigure}{.24\linewidth}
        \centering
        \includegraphics[width=1.0\textwidth]{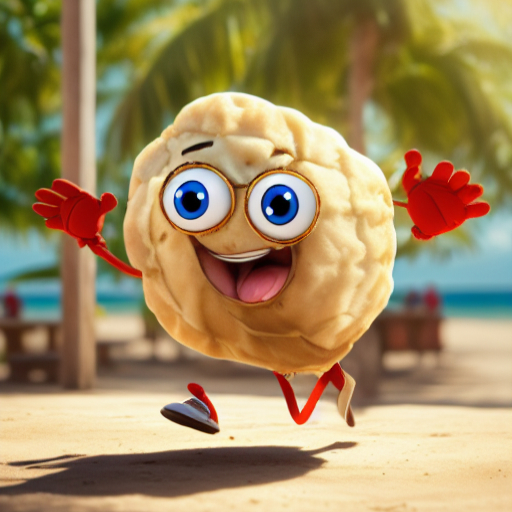}
        \caption{Result}
        \label{failure-d}
    \end{subfigure}
    \vspace{-3mm}\caption{Failure cases. Edits may negate each other.}\vspace{-6mm}
    \label{fig:failure}
\end{figure}

Similar to DragDiffusion, one limitation of our approach is the challenge in handling real images while preserving their original fidelity. Currently, we rely on methods like LoRA to maintain the integrity of real images during the editing process. While effective, this method can sometimes constrain the range and depth of the editing capabilities.
{Moreover, as shown in \cref{fig:failure}, our method sometimes fails in executing intended rotations, underlining the constraints of point-based editing in tasks needing a global perspective, such as novel view synthesis or full object rotation. Future research should aim at creating a universal adapter to maintain diffusion model fidelity and improve editing that requires integrated point understanding.}

\noindent\textbf{Acknowledgement.}
The work is supported by the Guangdong Natural Science Funds for Distinguished Young Scholar (No. 2023B1515020097), Singapore MOE Tier 1 Funds (No. MSS23C002), NRF Singapore under the AI Singapore Programme (No. AISG3-GV-2023-011), Guangzhou 2023 Key Research and Development Program on Science and Technology for Agriculture and Social Development (2023B03J1328), and the National Key Technologies Research and Development Program of China (2023YFD2400600).
%\vspace{-10mm}
\newpage 
{
    \small
    \bibliographystyle{ieeenat_fullname}
    \bibliography{main}
}


\end{document}